\documentclass[10pt,twocolumn,letterpaper]{article}

\usepackage[pagenumbers]{cvpr} 

\usepackage[table,dvipsnames]{xcolor}
\definecolor{lightgray}{gray}{0.9}

\usepackage{wrapfig}

\definecolor{cvprblue}{rgb}{0.21,0.49,0.74}
\usepackage[pagebackref,breaklinks,colorlinks,citecolor=cvprblue]{hyperref}
\usepackage{cite}
\usepackage[utf8]{inputenc} 
\usepackage[T1]{fontenc}    
\usepackage{hyperref}       
\usepackage{url}            
\usepackage{booktabs}       
\usepackage{amsfonts}       
\usepackage{nicefrac}       
\usepackage{microtype}      
\usepackage{amsmath}
\usepackage{amssymb}
\usepackage{graphicx}
\usepackage{algorithm}
\usepackage{algpseudocode}
\usepackage{enumitem}
\usepackage{multirow}
\usepackage{wrapfig}

\def\paperID{1478} 
\def\confName{CVPR}
\def\confYear{2024}

\title{Training Like a Medical Resident: Context-Prior Learning Toward Universal Medical Image Segmentation}

\author{Yunhe Gao\textsuperscript{1}
\and Zhuowei Li\textsuperscript{1}
\and Di Liu\textsuperscript{1}
\and Mu Zhou\textsuperscript{1}
\and Shaoting Zhang\textsuperscript{2}
\and Dimitris Metaxas\textsuperscript{1}\\
\textsuperscript{1}Rutgers University,\quad\quad\quad
\textsuperscript{2}Shanghai AI Laboratory
}

\begin{document}
\maketitle
\begin{abstract}

A major focus of clinical imaging workflow is disease diagnosis and management, leading to medical imaging datasets strongly tied to specific clinical objectives. This scenario has led to the prevailing practice of developing task-specific segmentation models, without gaining insights from widespread imaging cohorts. Inspired by the training program of medical radiology residents, we propose a shift towards universal medical image segmentation, a paradigm aiming to build medical image understanding foundation models by leveraging the diversity and commonality across clinical targets, body regions, and imaging modalities. Towards this goal, we develop \textbf{Hermes}, a novel context-prior learning approach to address the challenges of data heterogeneity and annotation differences in medical image segmentation. In a large collection of eleven diverse datasets (2,438 3D images) across five modalities (CT, PET, T1, T2 and cine MRI) and multiple body regions, we demonstrate the merit of the universal paradigm over the traditional paradigm on addressing multiple tasks within a single model. By exploiting the synergy across tasks, Hermes achieves state-of-the-art performance on all testing datasets and shows superior model scalability. Results on two additional datasets reveals Hermes' strong performance for transfer learning, incremental learning, and generalization to downstream tasks. Hermes's learned priors demonstrate an appealing trait to reflect the intricate relations among tasks and modalities, which aligns with the established anatomical and imaging principles in radiology. The code is available\footnote{https://github.com/yhygao/universal-medical-image-segmentation}.

\end{abstract}    
\vspace{-1em}
\section{Introduction}
\label{sec:intro}

Medical image segmentation methods generate accurate delineations of anatomical structures which are crucial for disease diagnosis~\cite{de2018clinically,shen2015multi} and treatment planning~\cite{nestle2005comparison,gao2021focusnetv2,gao2019focusnet}. To date, the prevailing paradigm for medical image segmentation promotes the development of separate models for specific medical objects (e.g., organs or tumors) and image modalities (e.g., CT or MR)~\cite{isensee2021nnu,gao2021utnet,liu2022transfusion,zhangli2022region,chang2022deeprecon,gao2019multi}. This paradigm is often constrained by the limited training data from the same domain, resulting in compromised model robustness and generalizability. Further scaling up data size for specific segmentation tasks is challenging due to the high cost of data acquisition, collection, and annotation~\cite{razzak2018deep,gao2021enabling}. Moreover, the current paradigm is unable to exploit relationships among medical imaging tasks that are critical for disease understanding. These hurdles together confine the capability and scalability of medical image segmentation models.

\begin{figure}[t]
\begin{center}
\includegraphics[width=0.47\textwidth]{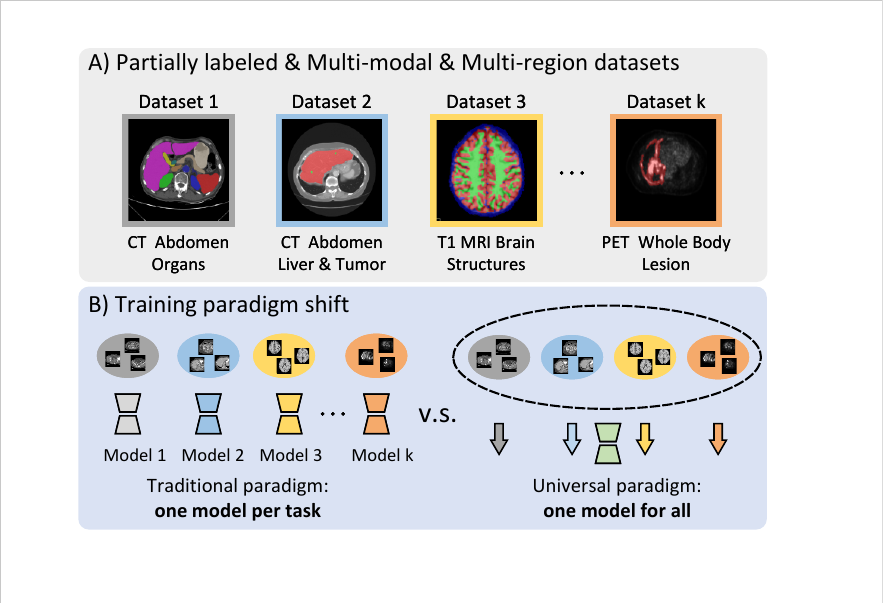}
\end{center}
\vspace{-1em}
\caption{A) Clinical diagnostic workflows typically focus on specific specialties and diseases, leading to the curation of image datasets that are partially annotated, multi-modal, and multi-regional. B) Traditional training paradigms involve training separate models for each segmentation task (or dataset). In contrast, we emphasize a universal medical image segmentation paradigm aiming at \textbf{one model for all}, leading to a robust and generalizable universal model for diverse tasks.}
\label{fig:fig1}
\vspace{-2em}
\end{figure}

Inspired by radiology residency programs~\cite{acgme2022,acgme2023number,sarkany2018running}, we recognize that radiologists' expertise arises from routine exposure to wide-ranging images across body regions, diseases, and imaging modalities. Despite the fact that the human body exhibits anatomical variability, it is fundamentally composed of various types of tissues and structures whose appearance in images are often statistically similar~\cite{scanlon2018essentials}. These tissues display specific visual characteristics under different imaging modalities~\cite{bankman2008handbook,hussain2022modern}. During the training process, residents acquire prior knowledge about the regions of interest (ROI) across imaging modalities for interpreting various types of patterns. Inspired by this observation, we prioritize a paradigm shift toward universal medical image segmentation, which seeks to harness the diversity and commonality of medical images, to build a comprehensive, unified segmentation model. However, addressing the heterogeneity among medical imaging tasks faces daunting challenges.

First, each image dataset has partial and incomplete ROI annotations due to their distinct needs and clinical objectives. As seen in Figure \ref{fig:fig1}, dataset 1 contains annotations for 15 abdominal organs, while dataset 2 is only annotated for the liver and tumor. Secondly, class definitions can vary depending on the clinical target. To illustrate this point, in dataset 1, the liver is annotated as a whole organ, whereas in dataset 2, the same liver region is divided into two categories of liver and tumor. Additionally, images from different datasets can exhibit significant statistical divergence due to factors such as imaging modalities and body regions (dataset 3 and k).

Given these challenges, intriguing questions remain to be fully explored. How can segmentation tasks with different target ROIs mutually benefit from each other? Is the underlying feature representation transferable across different body regions? Despite the visual difference in imaging modalities, how can a model discern and utilize the meaningful commonalities between them?

In this paper, we strive to answer the above questions. Our main contributions are:

\begin{itemize}[leftmargin=*]

\item By exploring the universal medical image segmentation paradigm, we show a single unified model can handle tasks across various ROIs, anatomical regions, and modalities.

\item We introduce a novel context-prior pool to learn two important types of prior knowledge, \textit{task} and \textit{modality}, directly from medical images. Different from using one-hot~\cite{zhang2021dodnet} or CLIP embeddings~\cite{liu2023clip} to inject task information, Hermes's learned priors are able to capture complex relations among tasks and modalities that are aligned with established anatomical and imaging principles.

\item Through a carefully curated eleven datasets, our systematic analysis reveals the strong capability of Hermes in accuracy and model scalability.

\item With two additional downstream datasets, we show Hermes's superior performance in transfer learning, incremental learning, and generalization, affirming the efficacy of the universal paradigm in acquiring robust and generalizable image representations.

\end{itemize}

\section{Related Work}
\label{sec:related_work}

\textbf{Partially labeled data problem.} 
Disease-specific clinical workflows often yield datasets with single-modality images and partial annotations tailored to specific clinical objectives. Early efforts aimed to combine multiple datasets by conditioning on labels~\cite{dmitriev2019learning}, regularizing organ size distribution~\cite{zhou2019prior}, or generating multi-organ pseudo labels for co-training~\cite{huang2020multi}. Recent approaches propose different conditioning methods to inject task information. DoDNet~\cite{zhang2021dodnet} embeds the task index with one-hot vectors as additional model inputs, but these one-hot task vectors are orthogonal with no task relations embedded. CLIP-driven universal model~\cite{liu2023clip} conditions on CLIP text embeddings, but the CLIP encoder, rarely trained with medical data, struggles with the semantics and relationships of complex medical concepts (Fig. \ref{fig:prior_analysis}, \ref{fig:mod_prior_analysis}). MultiTalent~\cite{ulrich2023multitalent} uses multiple task-specific heads. Moreover, DoDNet, CLIP-driven and MultiTalent limit their scope to a single body region or imaging modality. UniSeg~\cite{ye2023uniseg} introduces novel learnable task prompts but does not consider the modality information. In contrast, our work focuses on learning priors directly from a diverse array of medical data sources by addressing both task and modality heterogeneity simultaneously.

\noindent\textbf{Prior knowledge in medical image analysis.} Prior knowledge typically involves the understanding of anatomical structures or specifics of imaging modalities before further processing. Traditionally, atlas-based techniques~\cite{iglesias2015multi,cabezas2011review} and statistical shape models~\cite{cootes1995active}, leverage prior anatomical knowledge by aligning a predefined atlas or template to a patient's image. In deep learning, major methods embed prior knowledge into network architectures or training strategies. For example, integrating graphical models to embed spatial and anatomical knowledge~\cite{kamnitsas2017efficient,chen2019vertebrae}, or introducing a shape prior knowledge regularization loss through an autoencoder~\cite{oktay2017anatomically,gao2021focusnetv2}. Our work diverges by harnessing context-prior learning to learn task and modality knowledge directly from medical data, where this prior knowledge is injected into the backbone to enhance segmentation during inference.

\noindent\textbf{Universal image segmentation} aims to unify semantic, instance, and panoptic segmentation into one framework~\cite{zhang2021k,cheng2021per,cheng2022masked}. However, in 3D medical imaging, it is uncommon to find multiple instances of the same object. In addition, fully annotated medical datasets are rarely available in the community, making instance and panoptic segmentation highly unsuitable. As a result, universal medical image segmentation focuses on the semantic segmentation of medical objects. While sharing similarities, developing universal medical image segmentation possesses unique challenges, including the presence of partial annotation, conflicting class definitions, and heterogeneous medical-image contents~\cite{razzak2018deep}. 

\section{Method}
\label{sec:method}

\begin{figure*}[t]
\begin{center}
\includegraphics[width=0.8\textwidth]{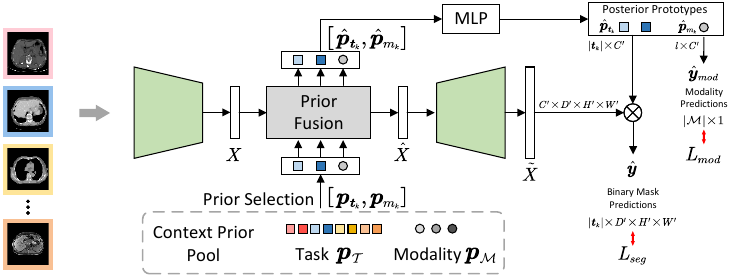}
\end{center}
\caption{Illustration of Hermes. A context-prior knowledge pool, including task and modality priors, is learned with the backbone. Through oracle-guided selection and combination of these priors, Hermes can address a variety of segmentation tasks and image modalities. }
\label{fig:framework}
\vspace{-1em}
\end{figure*}

\subsection{Preliminary}
\textbf{Problem definition.} We hereby define \textit{universal medical image segmentation} as the endeavor to train a universal model that learns from diverse medical datasets and performs various image segmentation tasks. Given a set of $K$ datasets, let $\mathcal{D}_k = \{(\boldsymbol{x}_{k}^{i}, \boldsymbol{y}_{k}^{i}, m_{k}, \boldsymbol{t}_{k})\}_{i=1}^{N_k}$ be the $k$-th dataset comprised of $N_k$ data pairs, where each data pair $(\boldsymbol{x}_{k}^{i},\boldsymbol{y}_{k}^{i})\in(\mathcal{X}_k \times \mathcal{Y}_k)$. $m_{k}$ denotes the imaging modality of $\mathcal{D}_k$ and $\boldsymbol{t}_{k}$ represents the tasks contained in $\mathcal{D}_k$, where each task represents the segmentation of a unique clinical target annotated in $\mathcal{Y}_k$, and $|\boldsymbol{t}_k|=|\mathcal{Y}_k|$. The collective dataset is given by $\mathcal{D} = \cup_{k=1}^{K} \mathcal{D}_k$, the corresponding collective tasks are $\mathcal{T}=\cup_{k=1}^K\boldsymbol{t}_k$, and the collective modalities are $\mathcal{M}=\cup_{k=1}^Km_k$. The objective is to train a single model $f_\theta: \mathcal{X} \rightarrow \mathcal{Y}$, parameterized by $\theta$, where $\mathcal{X}=\cup_{k=1}^K\mathcal{X}_k$ and $\mathcal{Y}=\cup_{k=1}^K\mathcal{Y}_k$.


\subsection{Oracle-guided context-prior learning}


In this section, we introduce Hermes, an oracle-guided context-prior learning approach, as depicted in Fig. \ref{fig:framework}. Drawing inspiration from radiology residency training, Hermes explicitly learns context-prior knowledge along with the segmentation backbone from diverse medical imaging sources. 
In line with radiologists' practice for image interpretation, Hermes uses the tasks to be processed $\boldsymbol{t}$,  and the imaging modality $m$ as guiding oracles during inference. These oracles guide the selection and combination of two types of priors - \textit{task} and \textit{modality} - to aid the backbone in addressing any specified tasks from the collective dataset. As a model-agnostic approach, Hermes can be seamlessly integrated with existing segmentation backbones, offering a versatile solution for universal medical image segmentation.

\noindent\textbf{Task context prior.} 
To learn from varied datasets and handle the incomplete annotation and potential conflict of class definition,  each task in $\mathcal{T}$ is formed as a binary segmentation task. For example, if ROIs in different datasets share the exact same definition, they are considered a single task, otherwise are treated as separate tasks. Each task is associated with a unique task prior, which is implemented as a learnable token. For the collective dataset $\mathcal{D}$ comprising $|\mathcal{T}|$ tasks, we initialize a task prior pool $\boldsymbol{p}_{\mathcal{T}}\in \mathbb{R}^{|\mathcal{T}|\times C}$, where $C$ denotes the token dimension.  Given a training image from $\mathcal{D}_k$, we use the task IDs $\boldsymbol{t}_{k}$ as an oracle to guide the selection of corresponding task priors $\boldsymbol{p}_{\boldsymbol{t}_k}\in \mathbb{R}^{|\boldsymbol{t}_k|\times C}$ from the pool, on which the model is conditioned to complete specified tasks. This design enables Hermes to flexibly select and combine task priors based on clinical objectives, accommodating a wide array of medical segmentation tasks.

\noindent\textbf{Modality context prior.} Medical images come with a variety of imaging modalities, each possessing distinct image attributes, intensity profiles, and noise patterns. To reduce the modeling difficulty, we introduce a modality prior for each modality. Similar to the task prior, we initialize a modality prior pool $\boldsymbol{p}_{\mathcal{M}} \in \mathbb{R}^{|\mathcal{M}| \times l \times C}$, where $l$ denotes the length of the token, and $C$ is the dimension of each token. We employ multiple tokens of length $l$ for each modality. When an image with modality $m_k$ is processed, we select the corresponding modality prior token  $\boldsymbol{p}_{m_k}\in \mathbb{R}^{l\times C}$ and concatenate it with the task prior tokens: $\boldsymbol{p}=[\boldsymbol{p}_{\boldsymbol{t}_k}, \boldsymbol{p}_{m_k}]\in \mathbb{R}^{(|\boldsymbol{t}_k|+l)\times C}$. 

\noindent\textbf{Conditioned segmentation.} Given an encoded image feature map $X\in\mathbb{R}^{C\times D\times H\times W}$ from the segmentation backbone, where $D, H, W$ represent the 3D feature map sizes, a prior fusion module is applied to adaptively fuse context-prior tokens and image features:
\begin{equation}
\label{eq:fusion}
    \boldsymbol{\hat{p}}, \hat{X}=Fusion(\boldsymbol{p}, X).
\end{equation}
Here, $\boldsymbol{\hat{p}}=[\boldsymbol{\hat{p}}_{\boldsymbol{t}_k}, \boldsymbol{\hat{p}}_{m_k}]$ represents the posterior tokens and and $\hat{X}$ is the prior-injected feature map. The prior fusion module can be implemented with the attention mechanism. For CNN backbones, we employ bi-directional cross-attention modules. For Transformer, we merge the prior tokens into the attention module (detailed in the supplementary).  The posterior tokens, processed through a multilayer perceptron (MLP), yield posterior prototypes serving as class classifiers. The predicted probability distribution for each task is computed via the inner product of these posterior task prototypes and image features, followed by a sigmoid function:
\begin{equation}
    \boldsymbol{\hat{y}}=p_{\theta}(y_k=1|\boldsymbol{x},\boldsymbol{p}_{t_k})=\sigma(\boldsymbol{\langle \hat{p}}_{t_k}, \tilde{X}\rangle),
\end{equation}
where $\boldsymbol{\hat{p}}_{t_k}\in \mathbb{R}^{|\boldsymbol{t}_k|\times C'}$ is the posterior prototypes (overload the $\boldsymbol{\hat{p}}_{t_k}$ in equation (\ref{eq:fusion})), $\tilde{X}\in\mathbb{R}^{C'\times D'\times H'\times W'}$ is the output feature map of the decoder, $\boldsymbol{\hat{y}}\in \mathbb{R}^{|\boldsymbol{t}_k|\times D'\times H'\times W'}$ is the predicted binary probability map for tasks $\boldsymbol{t}_k$. To help modality knowledge learning, we use auxiliary supervision on modality priors. Specifically, we predict image modality using posterior modality prototypes via global average pooling and a linear layer: $\boldsymbol{\hat{y}}_{mod}=\text{Linear}(\text{GAP}(\boldsymbol{\hat{p}}_{m_k}))$.

During training, Hermes learns class-specific representations for each task prior token and typical image features for each modality prior. In inference, context-priors and image feature maps undergo a bi-directional update, where priors assimilate instance-specific information from feature maps, and in turn, feature maps are augmented with learned prior knowledge. This interactive process is realized by attention weights, making Hermes a dynamic and adaptive model that enhances segmentation performance.

\noindent\textbf{Hierarchical modeling.} The effectiveness of hierarchical modeling is well-established in dense prediction tasks~\cite{lin2017feature,chen2018encoder}. We apply our context-prior approach hierarchically at multiple scales. The posterior prototypes from each scale are concatenated together and processed by an MLP as the final prototypes for segmentation. This design allows Hermes to learn prior knowledge across different scales, effectively merging multi-scale contextual information to improve the segmentation performance.

\begin{algorithm}[h]
\small
\caption{Hermes Training}
\label{alg:Hermes_training}
\begin{algorithmic}[1]
\State \textbf{Input:} Training dataset $\mathcal{D} = \cup_{k=1}^{K} \mathcal{D}_k$, where $\mathcal{D}_k = \{(\boldsymbol{x}_{k}^i, \boldsymbol{y}_{k}^i, m_{k}^i, \boldsymbol{t}_{k}^i)\}_{i=1}^{N_k}$. Randomly initialized segmentation backbone $f_{\theta}$, task and modality prior pool $\boldsymbol{p}_{\mathcal{T}}$ and $\boldsymbol{p}_{\mathcal{M}}$

\While {\textit{not converged}}
\For {j in [1, 2, \dots, batch size]}
    \State Randomly choose a training sample $\boldsymbol{x}, \boldsymbol{y}$ from $\mathcal{D}$
    \State Task and modality priors selection: $\boldsymbol{p}_{\boldsymbol{t}}$ and $\boldsymbol{p}_{m}$
    \State Prior concatenation: $\boldsymbol{p} = [\boldsymbol{p}_{\boldsymbol{t}},\ \boldsymbol{p}_{m}]$ 
\EndFor

\State Assemble training mini-batch $\mathcal{B}$  
\State Predict segmentation and modality: $\boldsymbol{\hat{y}}, \boldsymbol{\hat{y}}_{mod} =f_{\theta}(\boldsymbol{x}, \boldsymbol{p})$
\State Compute loss: $L_{seg} + \lambda L_{mod}$
\State Update  $f_{\theta}$, $\boldsymbol{p}_{\mathcal{T}}$ and $\boldsymbol{p}_{\mathcal{M}}$
\EndWhile

\end{algorithmic}
\end{algorithm}


\noindent\textbf{Training and losses.} Hermes employs a joint training strategy, using mixed batch training that incorporates samples from multiple datasets within each mini-batch (see Algorithm \ref{alg:Hermes_training}). The primary segmentation loss is a combination of binary cross-entropy and Dice loss: $L_{seg} = L_{bce}(\boldsymbol{y},\boldsymbol{\hat{y}}) + L_{dice}(\boldsymbol{y}, \boldsymbol{\hat{y}})$, The auxiliary modality loss uses cross-entropy loss on the predicted modality $\boldsymbol{\hat{y}}_{mod}$ and ground truth modality $\boldsymbol{m}$: $L_{mod}=L_{ce}(\boldsymbol{\hat{y}}_{mod}, m)$. The backbone and priors are jointly optimized to synergize representation learning and context-prior knowledge learning.

\section{Results}
\label{sec:results}

\subsection{Experiments setup}
\textbf{Dataset selection and experiment design.} Our collective training dataset, detailed in Table \ref{tab:dataset}, comprises eleven public datasets, chosen for their variety in body regions, modalities, and clinical targets. We assess model scalability by exploring the effect of model size on performance in both traditional and universal segmentation approaches. In addition, we further evaluate Hermes' capability in downstream tasks under the settings of transfer learning, incremental learning and generalization with two additional datasets. Finally, we show the priors learned by Hermes are able to accurately capture the complex relationships between tasks and modalities.

\noindent\textbf{Standardized preprocessing.} Training with diverse and heterogeneous data is a non-trivial problem. We implement a standardized preprocessing pipeline for all datasets. Firstly, we align all images to the same coordinate system and resample the spacing to a uniform $1.5\times 1.5\times 1.5\ mm$. Subsequently, we normalize image intensities; for CT data, we employ a clipping window of [-990, 500], while for MR and PET data, we clip at the 2nd and 98th percentiles of the intensity distribution. Finally, we conduct z-score normalization on each image, ensuring that all data exhibits a zero mean and unit standard deviation.

\noindent\textbf{Implementation details.} We implement Hermes using PyTorch and train the model with a batch size of 16 over 200 epochs.  We use the LAMB~\cite{you2019large} optimizer with a learning rate of 0.002 and an exponential learning rate decay. Data augmentations, including random cropping, rotation, scaling, brightness, contrast, and gamma perturbation, are applied on the fly during training. We use a patch size of $128\times 128\times 128$ for 3D training. The context-prior learning method is applied at the scales of $4\times$, $8\times$, and $16\times$ down-sampling. Given the variation in annotation styles across datasets, Classes with identical names across datasets are considered unique ROIs, except in AMOS CT and AMOS MR, resulting in 71 task prior tokens. Different MRI sequences are treated as distinct modalities, leading to a total of five modalities (CT, T1 MRI, T2 MRI, cineMRI, and PET) in the modality prior pool, with the length $l$ set to 10. We set $\lambda=0.001$ for the auxiliary modality loss as it is much easier than segmentation. The details of the datasets and train/val/test split are available in the supplementary.

\begin{table}
  \caption{Datasets statistics. The upper datasets are for upstream training and analysis. The bottom two datasets are for downstream tasks on transfer learning, incremental learning, and generalization.}
  \label{tab:dataset}
  \centering
   \scriptsize
   \setlength{\tabcolsep}{1mm}{
  \begin{tabular}{ccccccc}
    \toprule
    Dataset         & Body Region   & Modality  & Clinical Target   & \#Cls &   Size     \\
    \midrule
    BCV~\cite{bcv}     & Abdomen       & CT        & Organs            & 13        & 30    \\
    LiTS~\cite{bilic2019liver}     & Abdomen       & CT        & Liver \& Tumor    & 2         & 131   \\
    KiTS~\cite{heller2019kits19}     & Abdomen       & CT        & Kidney \& Tumor    & 2         & 210   \\
    AMOS CT~\cite{ji2022amos}  & Abdomen       & CT        & Organs            & 15        & 300   \\
    SS T~\cite{structseg}& Thorax        & CT        & Organs            & 6         & 50    \\
    SS H~\cite{structseg} & Head \& Neck & CT   & Organs & 22 & 50\\
    AMOS MR~\cite{ji2022amos}  & Abdomen       & MRI       & Organs            & 13        & 60    \\
    CHAOS~\cite{CHAOS2021}    & Abdomen       & T1 \& T2 MRI & Organs         & 4         & 60    \\
    M\&Ms~\cite{campello2021multi} & Cardiac & cineMRI & Structures & 3 & 320 \\
    DLBS~\cite{rodrigue2012beta} & Brain & T1 MRI & Structures & 3 & 213 \\
    AutoPET~\cite{gatidis2022whole} & Whole body & PET & Lesions & 1 & 1014 \\
    \midrule
    SegTHOR~\cite{lambert2020segthor}  & Thorax        & CT        & Organs            & 3         & 40    \\
    MSD Pancreas~\cite{antonelli2022medical}& Abdomen       & CT        & Pancreas \& Tumor    & 2         & 281 \\

    \bottomrule
  \end{tabular}}
\end{table}

\subsection{Results}
\begin{table*}
  \caption{Universal segmentation results measured with Dice score (\%). The upper table presents models with CNN backbones. The lower table shows models with Transformer backbones. Hermes-R denotes our method using ResUNet backbone, while Hermes-M uses MedFormer backbone.}
  \label{tab:universal_results}
  \centering
  \footnotesize
  \setlength{\tabcolsep}{1.5mm}{
  \begin{tabular}{c|ccccccccccccc}
    \toprule
    Paradigm    & Model    & BCV    & SS T & SS H & LiTS T    & KiTS T    & AMOS CT   & AMOS MR   & CHAOS & M\&Ms &  AutoPET & DLBS &  AVG       \\
    \hline
    Traditional & nnUNet~\cite{isensee2021nnu}  & 84.23	&88.53 & 78.17	&64.91	&81.72	&88.79	&85.49	&91.34	&85.65 & 65.43 & 94.22 & 82.59\\
    Traditional &  ResUNet & 84.36	&88.59 & 78.12	&64.87	&81.89	&88.97	&85.43	&91.41	&85.74 & 65.52 & 94.31 & 82.65  \\
    SSL & DeSD~\cite{ye2022desd} & 83.62 & 88.11 & 76.56 & 64.43 & 82.52 & 86.36 & 82.56 & 91.55 & 86.46 & 69.02 & 86.81 & 81.64 \\
    \rowcolor{lightgray}
    Universal  &  Hermes-R & \bf85.98	&\bf89.50 & \bf 80.62 	&\bf67.49	&\bf85.46	&\bf89.63	&\bf86.78	&\bf92.01	&\bf 86.94 & \bf 73.69 & \bf 96.21 & \bf 84.93 \\\midrule

    Traditional  &  SwinUNETR~\cite{hatamizadeh2022swin} &83.32  & 88.36 & 72.21  & 64.82 & 74.32 & 88.29 & 83.97 & 88.34 & 83.28 & 64.39 & 92.01 & 80.32 \\
    Traditional   &  MedFormer~\cite{gao2022data} & 84.61 & 89.04 & 78.71 & 66.24 & 82.09 & 89.45 & 85.58 & 91.85 & 86.02 & 66.01 & 95.13 & 83.39 \\
    SSL & UniMiSS~\cite{xie2022unimiss} & 84.97 & 88.29 & 77.41 & 63.94 & 61.21 & 85.82 & 83.51 & 91.35 & 85.75 & 60.32 & 95.35 & 80.64\\
    \rowcolor{lightgray}
    Universal   & Hermes-M &\bf 86.29	&\bf 89.61 & \bf 81.19	&\bf 68.32	&\bf 85.98	&\bf 89.98	&\bf 87.20	&\bf 92.22	&\bf87.02 & \bf 74.04 & \bf 96.54 & \bf 85.28 \\

    \bottomrule
  \end{tabular}}
  \vspace{-1.4em}
\end{table*}

We start initial experiments under the traditional training paradigm by training individual models for each dataset. We use several representative medical image segmentation backbones, including CNN models nnUNet~\cite{isensee2021nnu}, UNet~\cite{ronneberger2015u} with residual building blocks~\cite{he2016deep}: ResUNet, and Transformer models SwinUNETR~\cite{hatamizadeh2022swin} and MedFormer~\cite{gao2022data}. From Table \ref{tab:universal_results}, ResUNet and MedFormer demonstrate slightly better results under the traditional paradigm. Therefore, in subsequent experiments under the universal paradigm, we employ ResUNet and MedFormer as representative backbones for CNN and Transformer to implement Hermes, i.e. Hermes-R and Hermes-M. We also compare against large-scale self-supervised pretraining and finetuning (SSL) methods: DeSD \cite{ye2022desd} (pretrained on 10,594 CT scans) and UniMiss \cite{xie2022unimiss} (pretrained with 5,022 CT scans and 108,948 2D images).


\noindent\textbf{Results on target ROIs, body regions, and image modalities.} From Table \ref{tab:universal_results}, we see that: \textbf{First}, a mutual enhancement in performance is observed across different target ROIs. Specifically, Hermes-R shows significant gains over ResUNet in tumor and lesion segmentation in the LiTS and KiTS datasets, increasing by 2.62\% and 3.57\%, respectively. Across the other ROIs in eleven datasets, Hermes consistently improves results, with Hermes-R outperforming ResUNet by 2.28\%, and Hermes-M surpassing MedFormer by 1.89\%. \textbf{Second}, there's a clear transferability of underlying representations between different body regions. For example, in thoracic and head\&neck regions represented by SS T and SS H datasets, Hermes-R enhances segmentation in thoracic targets by 0.91\% in SS T and by 2.5\% in SS H compared to ResUNet. \textbf{Third}, the commonality between imaging modalities can be harnessed to enhance performance, even though they possess entirely different visual characteristics. For instance, despite the inherent image differences between PET to other MRI and CT datasets, Hermes-R has a remarkable improvement of 8.17\% with the universal paradigm. \textbf{Lastly}, although DeSD and UniMiss are pretrained on much larger datasets than Hermes, they do not transfer well to different tasks, especially cross-modalities. For example, DeSD is notably worse in the DLBS with brain T1 MRI images, and UniMiss underperforms in AutoPET.

\begin{figure*}[t]
\begin{center}
\includegraphics[width=1\textwidth]{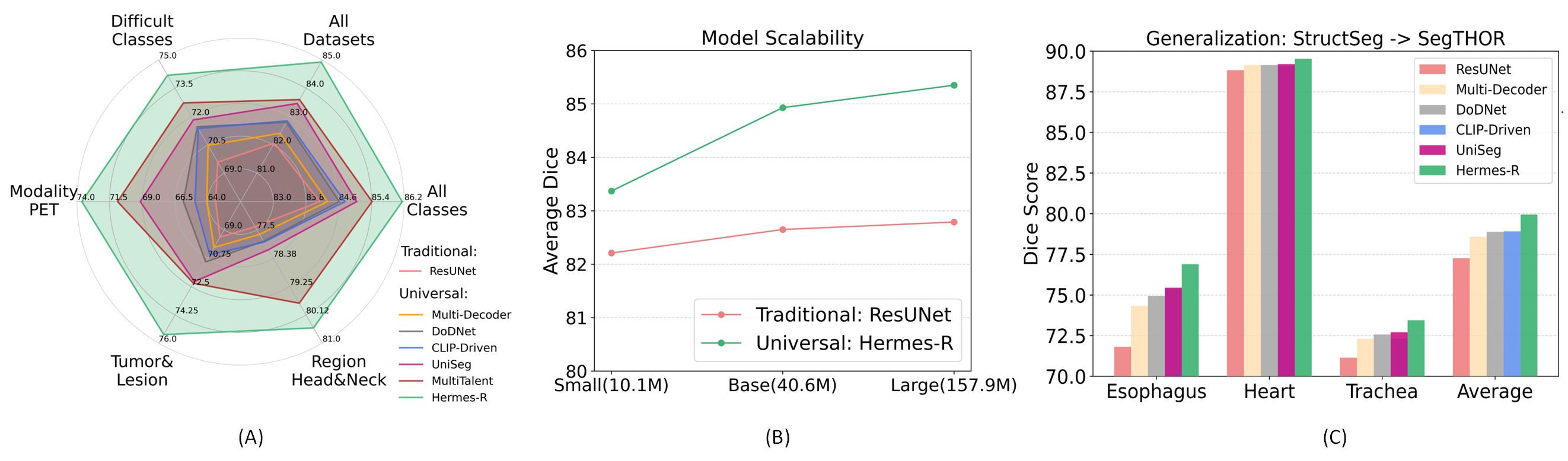}
\end{center}
\vspace{-2em}
\caption{(A) Comparison with other SOTA methods. ROIs with Dice scores lower than 80 under the traditional paradigm are defined as 'difficult classes'. (B) Model scalability analysis. We scale ResUNet down and up to three variants: ResUNet-Small (10.1M), ResUNet-Base (40.6M), and ResUNet-Large (157.9M), and the same for Hermes. All other experiments use ResUNet-Base as the backbone unless specified. (C) Generalization from StructSeg to SegTHOR.}
\label{fig:plot}
\vspace{-1em}
\end{figure*}

\noindent\textbf{Comparison with other methods under the universal paradigm.} We adapt and extend several previous methods to fit the universal paradigm. Multi-decoder~\cite{chen2019med3d} uses a shared encoder and separate decoders per dataset. DoDNet~\cite{zhang2021dodnet} was originally designed for organ and tumor segmentation in CT images with one-hot task vector embeddings; we broaden it into multi-modality segmentation. CLIP-Driven~\cite{liu2023clip} was intended for partially labeled CT image analysis using CLIP text encoder; we adapt it by including modality information in its text prompt: "\textit{A CT/MRI/PET of a [CLS].}" MultiTalent~\cite{ulrich2023multitalent} uses multiple segmentation heads. UniSeg~\cite{ye2023uniseg} uses learnable task prompts. All methods were implemented using the same ResUNet backbone for fairness.

Figure \ref{fig:plot} (A) presents a comprehensive comparison of traditional and universal paradigms from six perspectives, detailed numbers are in supplementary. All universal-trained methods demonstrate consistent improvement over the traditional paradigm, particularly for difficult classes, underscoring the universal paradigm's potency in deriving robust representations. Within the universal paradigm, Hermes outperforms other methods in all six perspectives. For example, Multi-Decoder's scalability is limited by the linear growth of decoders with dataset quantity, and it only allows knowledge sharing at the encoder level. DoDNet's one-hot task embedding is incapable of capturing inter-task relationships. CLIP-Driven, despite incorporating knowledge from text prompts, can't effectively encode discriminative information for medical tasks, as its CLIP text encoder lacks training on medical texts, leading to subpar performance, especially for tumors\&lesions. 
MultiTalent demonstrates good performance with task-specific segmentation heads, yet it does not reach the effectiveness of Hermes because it lacks the integration of prior knowledge during inference. Although UniSeg incorporates learnable task prompt maps, it overlooks modality information and learns prompts at the lowest resolution, missing critical multi-scale information. In contrast, Hermes possesses two key advances: (1) It utilizes versatile task and modality context priors, learned directly from medical data, which accurately capture the nuances of task and modality knowledge, as demonstrated in Fig. \ref{fig:prior_analysis} and \ref{fig:mod_prior_analysis}. (2) It employs hierarchical modeling to help multi-scale learning of prior knowledge.

\noindent\textbf{Model scalability.}  In Figure \ref{fig:plot} (B), we find that, in contrast to neural scaling law~\cite{kaplan2020scaling,he2016deep,dosovitskiy2020image,liu2021swin}, increasing the model scale yields marginal performance gain in the traditional paradigm due to limited size and diversity of individual dataset. The limited data from the same distribution cannot fully utilize the larger model's capacity, potentially leading to overfitting. On the contrary, the universal paradigm shifts this dynamic, with an increased backbone scale for Hermes leading to a remarkable performance gain. This finding suggests the proposed universal paradigm's ability to harness the potential of larger models with diverse data and tasks. This approach paves the way for training large-scale models in medical imaging, a feat that is challenging under traditional training paradigms.

\subsection{Downstream tasks}
In the universal paradigm, the upstream multi-task training encourages the models to learn robust and generalizable representations across various tasks. We extend to perform an analysis of downstream tasks in the following section.

\begin{table*}
  \caption{Transfer and incremental learning on the MSD Pancreas \& Tumor~\cite{antonelli2022medical} dataset}
  \label{tab:transfer}
  \centering
  \footnotesize
  \setlength{\tabcolsep}{3mm}{
  \begin{tabular}{c|c|cc|cc|cc|cc}
    \toprule
    Setting    & Model    & \multicolumn{2}{c}{1\%}  & \multicolumn{2}{c}{10\%}  & \multicolumn{2}{c}{50\%}  & \multicolumn{2}{c}{100\%}      \\\hline
            &           &  Pan & Tumor & Pan & Tumor & Pan & Tumor & Pan & Tumor \\
    Scratch & ResUNet	&44.60	&7.67	&74.47	&23.90 &78.89	&44.52 &80.45	&51.06\\\hline
    \multirow{7}{*}{Transfer}  &  ResUNet (AMOS CT) &56.08	&8.31 &77.15	&25.53  &80.53	&46.16	&81.23	&52.21\\
                        ~ & ResUNet (KiTS) &52.68	&9.28 &75.11	&27.33	&79.07	&45.72	&79.23	&50.64 \\
                        ~ & DeSD~\cite{ye2022desd} (10,594 CT) & 67.82 & 13.89 & 78.11 & 35.82 & 80.95 & 50.23 & 81.97 & 59.11 \\\cline{2-10}
                        ~ & DoDNet~\cite{zhang2021dodnet} & 66.62	& 11.97 & 76.83	& 31.92 & 80.82	& 47.79 & 81.41	& 53.62\\
                        ~ & CLIP-Driven~\cite{liu2023clip} &   67.95	& 12.12       & 77.49	&32.37   &80.92	   &48.92   &81.45	   &54.71   \\
                        ~ & UniSeg~\cite{ye2023uniseg} & 69.05 & 12.35 & 77.33 & 33.87 & 80.93 & 49.63 & 81.96 & 55.58 \\
                        ~ & \cellcolor{lightgray}Hermes-R & \cellcolor{lightgray}\bf 72.71   & \cellcolor{lightgray}\bf 16.73  & \cellcolor{lightgray}\bf 79.12	& \cellcolor{lightgray}\bf 44.31  & \cellcolor{lightgray}\bf 81.14     & \cellcolor{lightgray}\bf 55.31    & \cellcolor{lightgray}\bf 82.73	& \cellcolor{lightgray}\bf 61.41\\\hline
    \multirow{4}{*}{Incremental} & DoDNet~\cite{zhang2021dodnet} &64.36	&7.91			&65.78	&14.39	&69.28	&21.97	&69.87	&22.32 \\  
                        ~ & CLIP-Driven~\cite{liu2023clip} & 71.52	&10.79 &75.41	&23.22	&76.25	&29.02	&74.72	&30.99\\
                        ~ & UniSeg~\cite{ye2023uniseg} & 71.68 & 12.31 & 72.96  & 19.82 & 76.95 & 27.02 & 74.40 & 28.22\\
                        ~ & \cellcolor{lightgray}Hermes-R & \cellcolor{lightgray}\bf 72.69	& \cellcolor{lightgray}\bf15.52	& \cellcolor{lightgray}\bf76.89	&\cellcolor{lightgray}\bf28.61	&\cellcolor{lightgray}\bf79.44	&\cellcolor{lightgray}\bf43.12	&\cellcolor{lightgray}\bf79.98	&\cellcolor{lightgray}\bf 47.12\\
                       
    \bottomrule
  \end{tabular}}
\vspace{-1.4em}
\end{table*}

\noindent\textbf{Transfer learning.} We perform evaluation on the challenging MSD pancreas \& tumor dataset~\cite{antonelli2022medical}, divided into training (214 samples), validation (10 samples), and testing (57 samples) sets. To assess the influence of downstream data volume, models are fine-tuned on 1\%, 10\%, 50\%, and 100\% of the training samples. We report average Dice over 5 runs to reduce variability from the training samples. 

Table \ref{tab:transfer} compares the efficacy of different transfering methods. Traditional single-dataset pretraining with AMOS CT (15 organs) or KiTS (kidney \& tumor) shows improved performance with a small amount of downstream data (1\%-10\%), but these gains saturate as the volume of downstream task data increases (50\%-100\%). Notably, KiTS pretraining even lags behind training from scratch when using the full data. In contrast, the improvements offered by the universal paradigm, including DoDNet, CLIP-Driven, and Hermes, consistently outperform single-dataset transfer. Despite DeSD's self-supervised pretraining on a larger dataset, Hermes exhibits superior transferability across all data volumes. Hermes particularly excels in both constrained (1\%-10\%) and abundant (50\%-100\%) data scenarios, demonstrating the value of upstream task diversity in enhancing transfer learning for downstream medical image segmentation tasks.

\noindent\textbf{Incremental learning.} We test model performance in a more challenging scenario that requires the model to retain knowledge from previous tasks while learning new ones. For all four compared methods, incremental learning is accomplished by keeping the backbone fixed and finetuning only the new task conditions. Table \ref{tab:transfer} shows that Hermes excels in this setting, particularly under limited data conditions (1\%-10\%), even outperforming transfer learning results from AMOS CT and KiTS. With ample data, Hermes maintains competitive performance comparable to that of full model tuning. This capability highlights the strength of the backbone's representation learning, fostered during diverse upstream universal training, and shows Hermes' adaptability to new downstream medical tasks.

\noindent\textbf{Generalization.} We directly apply models trained on the StructSeg to the SegTHOR datasets, both of which contain thoracic CT scans, and evaluate the performance on three overlapping categories as seen in Figure \ref{fig:plot} (C). Compared to ResUNet based on the single-dataset training, all methods trained under the universal paradigm show better generalization, with Hermes leading the performance. This outcome suggests that even with StructSeg being the only thoracic data in our upstream universal training, the inclusion of diverse data enhances potential generalization capabilities.

\subsection{Analysis}

\begin{figure}[t]
\begin{center}
\includegraphics[width=0.48\textwidth]{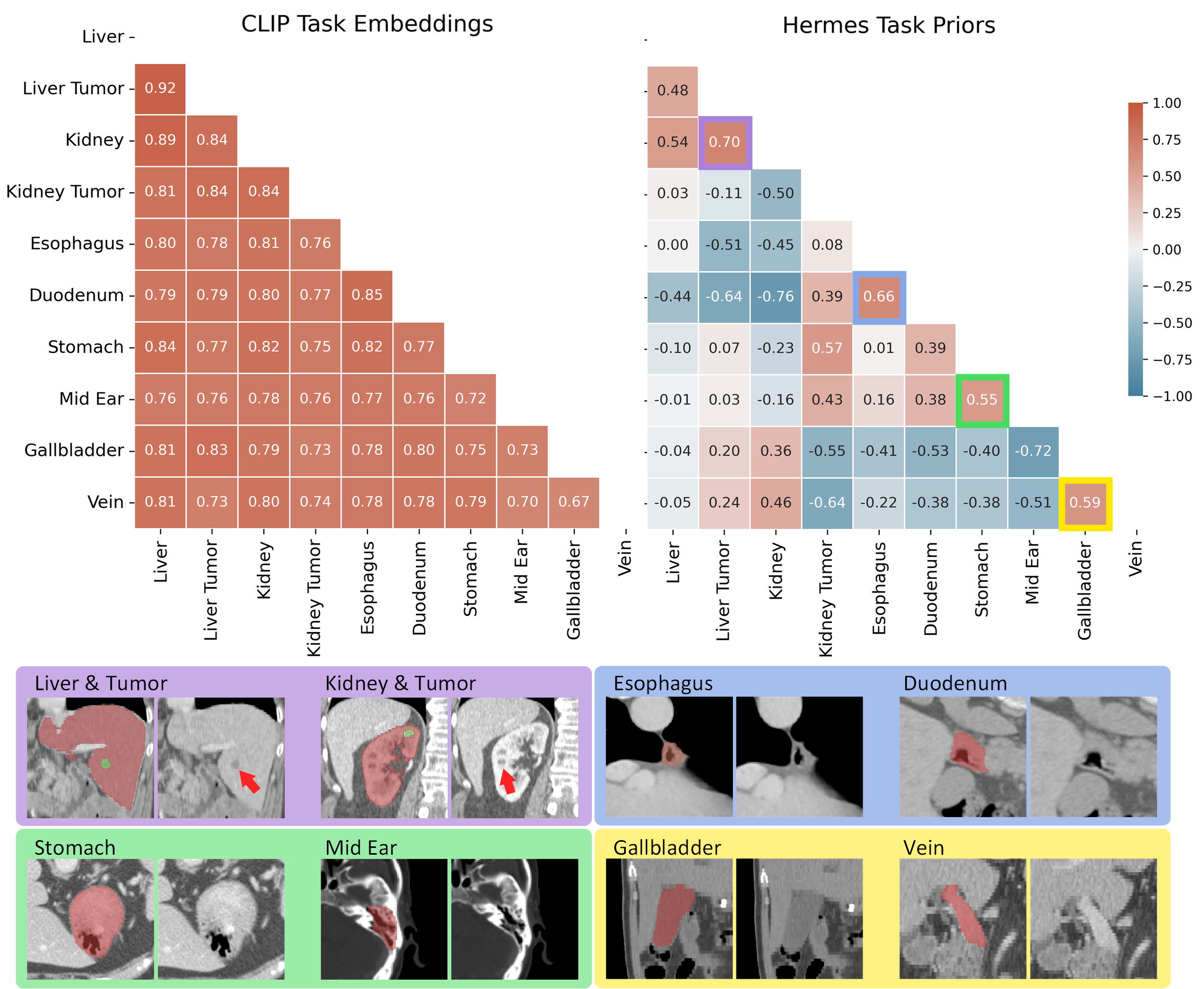}
\end{center}
\vspace{-1em}
\caption{Upper: Cosine similarity of Hermes's learned task priors and CLIP's task embeddings. Lower: Example structures that have high similarity. Hermes's priors are learned directly from medical data and are able to capture intricate relationships among tasks, while CLIP's embeddings tend to encode all objects into similar embeddings, resulting in a loss of discriminative details. }
\label{fig:prior_analysis}
\vspace{-1.5em}
\end{figure}

\begin{figure}[tb]
\centering
\includegraphics[width=0.43\textwidth]{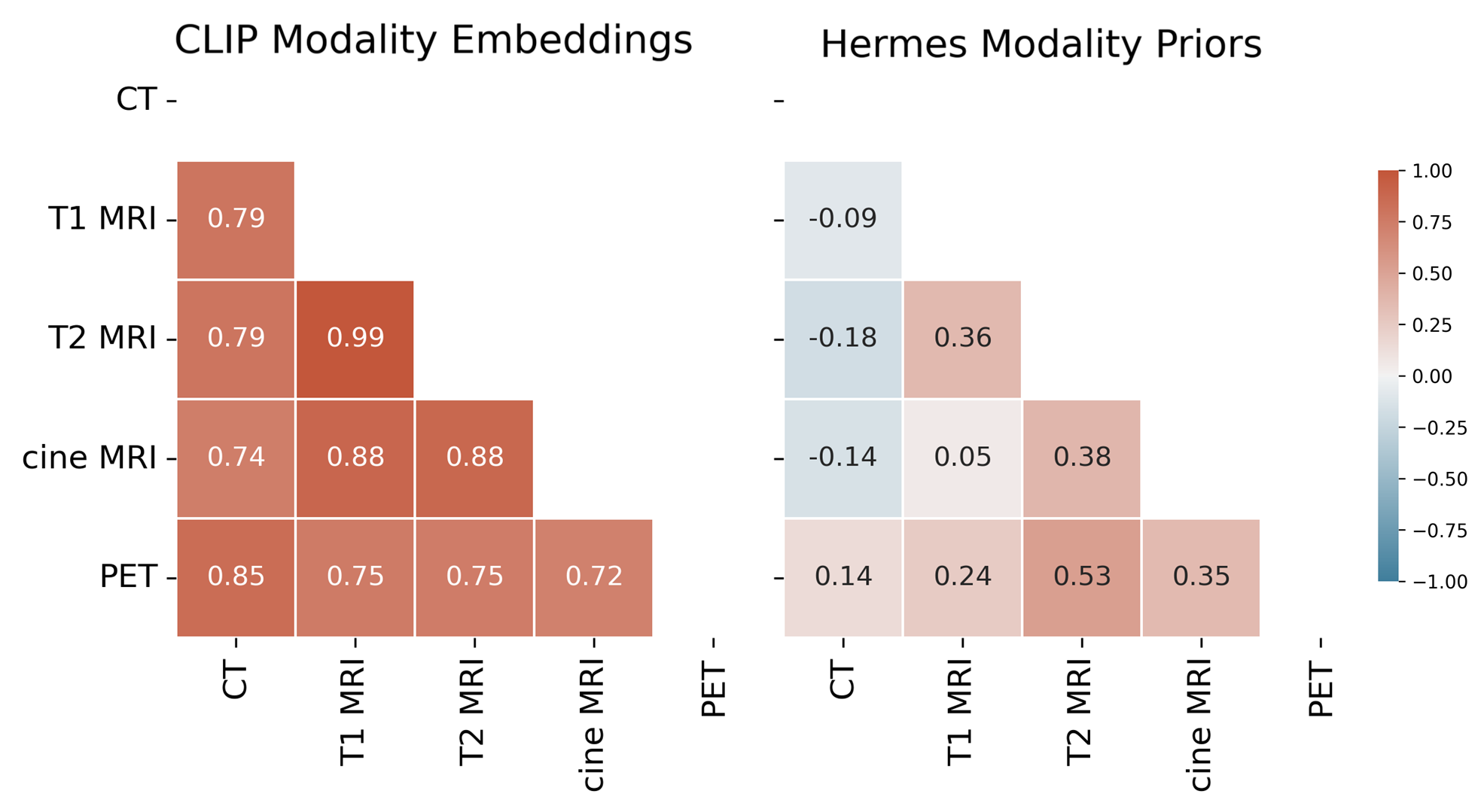}
\vspace{-1em}
\caption{Cosine similarity of Hermes's learned modality priors and CLIP's modality embeddings. The learned modality prior knowledge of Hermes is consistent with imaging principles.}
\label{fig:mod_prior_analysis}
\vspace{-2em}
\end{figure}

\noindent\textbf{Learned prior analysis.} We illustrate how Hermes adeptly learns task and modality context priors that are consistent with the established anatomical and imaging principles. In Fig. \ref{fig:prior_analysis}, we show the cosine similarity between task priors learned with Hermes and the one of CLIP text embeddings used in \cite{liu2023clip}. Note that this is pure prior knowledge without considering any image features. The CLIP text encoder, rarely trained with medical data, tends to encode all medical objects with high cosine similarity, offering limited prior knowledge about tasks. In contrast, \textbf{Hermes excels in reflecting the intricate relationships among tasks by learning directly from medical images.} This is evident in how objects with similar visual features show high similarity, and vice versa. For example, liver tumors display a notable similarity with kidneys, attributable to their shared visual features with the renal medulla or pelvis. Also, the duodenum and esophagus, being parts of the digestive system, demonstrate similarity in their tubular structures filled with air. Intriguingly, despite their significant anatomical and functional differences, the middle ear and stomach are shown to have comparable densities in CT scans, attributed to their air-filled hollows. In contrast, fluid-filled structures like the gallbladder and veins demonstrate higher similarity to each other but differ markedly from hollow organs such as the esophagus, stomach, or middle ear. \textbf{Hermes also captures accurate prior knowledge about imaging modalities}. As illustrated in Fig. \ref{fig:mod_prior_analysis}, Hermes effectively discerns distinct characteristics of various imaging modalities. Hermes identifies CT as quite distinct from other modalities like MRI and PET, due to CT's unique imaging principle based on X-ray absorption. Meanwhile, T2 MRI and cine MRI, which both highlight fluid content, align closely with each other  while PET, focused on metabolic activities, shows greater similarity to T2 and cine MRI than to T1 MRI and CT. Further elaboration on these findings is provided in the supplementary. These findings emphasize the capability of Hermes to learn meaningful and accurate prior knowledge from diverse medical images.


\begin{table}
    \small
    \caption{Ablation study}  \label{tab:ablation}
    \begin{tabular}{lc} 
    \toprule
    Design & Avg Dice\\\midrule
    Task prior only single scale & 81.72\\
    Task prior only + hierarchy & 82.98\\
    Task prior only - unidirectional fusion & 82.04\\
    Task + modality priors &  83.21\\
    Task + modality priors + modality auxiliary loss & 83.37\\
    \bottomrule
    \end{tabular}
    \vspace{-1em}
\end{table}
\noindent\textbf{Ablation study.} We conducted ablation studies using the ResUNet-Small backbone with Hermes to assess the impact of various configurations. We start with Hermes with only the task prior applied on the $16\times$ down-sampled scale, which is the most basic configuration of our method. Hierarchically integrating the task priors at multiple scales (e.g., $4\times$, $8\times$, and $16\times$) enhances the average Dice from 81.72\% to 82.98\%. In the prior fusion module, we employ a bidirectional attention mechanism to concurrently update both the prior and image features. To evaluate the effectiveness of this design, we alter the bidirectional attention to a unidirectional cross-attention. This modification only allows the prior to aggregate information while keeping the image features unchanged, similar to dynamic kernel approaches, e.g.~\cite{zhang2021dodnet,zhang2021k}. This change leads to a decreased Dice from 82.98\% to 82.04\%. Incorporating modality priors further improved the model's capacity to handle complex, heterogeneous multi-modality data, increasing the Dice score from 83.11\%. Adding an auxiliary loss can slightly boost the performance to 83.37\%.

\section{Discussion and Conclusion}
\label{sec:discussion}


\textbf{Implications of Hermes.} 
Our work introduces a holistic and universal paradigm in medical image analysis, leveraging the diversity and commonality among medical tasks in contrast to developing task-specific models~\cite{isensee2021nnu,gao2022data}. Despite exhibiting variability, different anatomical structures can share similar visual cues and enhance each other, evidenced by our results on in Table \ref{tab:universal_results}. Moreover, the interdisciplinary nature of our paradigm makes it possible to explore into deeper, more complex relationships within medical imagery, as initiated in our analysis (Figs. \ref{fig:prior_analysis} and \ref{fig:mod_prior_analysis}).

Additionally, the universal paradigm is a cost-efficient approach to scaling medical imaging usage. Given the challenges in assembling large, fully-annotated datasets, Hermes adeptly incorporates diverse, partially annotated datasets, addressing a broad spectrum of clinical targets. This flexibility enables the training of Hermes on a wide range of datasets with varying annotations, distinguishing it from conventional task-specific models and making it also an effective pretraining method with superior transferability. Such adaptability not only makes Hermes a practical asset in medical imaging but also signals a shift towards more versatile, data-inclusive approaches in the field.



\noindent\textbf{Connection with natural image foundation model.} Advances in foundational segmentation models (e.g., Segment-Anything Model (SAM)~\cite{kirillov2023segment}) have demonstrated notable progress for general vision tasks. Yet general-purpose models struggle to transfer to medical imaging tasks. For instance, a benchmark study~\cite{he2023accuracy} comparing SAM against 12 medical imaging tasks revealed that SAM's Dice scores consistently fell behind those of UNet by margins ranging from 0.1 to 0.5, and even reaching up to 0.6-0.7. Efforts of finetuning SAM are considerable to enhance its medical-image performance (e.g. MedSAM~\cite{ma2023segment}). Despite improvements over SAM, MedSAM's performance remains inferior to that of our 3D ResUNet baseline, even when the latter is trained from scratch with a single dataset. For example, on CT liver: ResUNet: 97.64 vs. MedSAM: 91.42, and CT pancreas: ResUNet: 80.45 vs. MedSAM: 76.76. The underperformance of SAM in medical imaging tasks may originate from: (1) the substantial domain gap~\cite{raghu2019transfusion} and (2) its inherent 2D design is ill-equipped to harness the 3D information that is crucial to medical image tasks. Therefore, building a foundation model for 3D medical images should be deeply rooted in 3D medical imaging itself. The proposed universal segmentation paradigm, along with Hermes, offers a versatile means for utilizing diverse, large-scale medical image datasets, opening up perspectives for the development of foundational models in medical imaging.

\noindent\textbf{Limitations.} While our model extends to leverages the breadth of partially labeled data, its performance can be potentially improved by addressing missing labels with self-supervised~\cite{he2022masked} or semi-supervised learning~\cite{tarvainen2017mean,sohn2020fixmatch}. Integrating more types of medical prior knowledge into model inference presents an intriguing topic for future investigation.

\noindent\textbf{Conclusion.} We introduce a universal medical image segmentation paradigm that learns generalizable and transferable representations from diverse medical image tasks, spanning various targets, body regions, and modalities. Following this paradigm, we propose a context-prior learning approach, Hermes, to tackle the challenges of annotation and modality heterogeneity. Hermes can handle multiple tasks by harnessing their synergy via our proposed task priors. Extensive experiments with Hermes underscore the superiority of this universal paradigm in both upstream and downstream tasks. Exploratory analysis of the learned priors shows intriguing relations among tasks and modalities, resonating with the anatomical and imaging principles in radiology. 

\noindent\textbf{Acknowledgement.} This research has been partially funded by research grants to D. Metaxas through NSF: 2310966, 2235405, 2212301, 2003874, and FA9550-23-1-0417 and NIH 2R01HL127661.

{
    \small
    \bibliographystyle{ieeenat_fullname}
    \bibliography{reference.bib}

\begin{thebibliography}{67}
\providecommand{\natexlab}[1]{#1}
\providecommand{\url}[1]{\texttt{#1}}
\expandafter\ifx\csname urlstyle\endcsname\relax
  \providecommand{\doi}[1]{doi: #1}\else
  \providecommand{\doi}{doi: \begingroup \urlstyle{rm}\Url}\fi

\bibitem[acg(2022)]{acgme2022}
Acgme program requirements for graduate medical education in diagnostic radiology.
\newblock Online, 2022.
\newblock Accessed: 2023/11/08.

\bibitem[acg(2023)]{acgme2023number}
Diagnostic radiology case log categories and required minimum numbers review committee for radiology.
\newblock Online, 2023.
\newblock Accessed: 2023/11/08.

\bibitem[Antonelli et~al.(2022)Antonelli, Reinke, Bakas, Farahani, Kopp-Schneider, Landman, Litjens, Menze, Ronneberger, Summers, et~al.]{antonelli2022medical}
Michela Antonelli, Annika Reinke, Spyridon Bakas, Keyvan Farahani, Annette Kopp-Schneider, Bennett~A Landman, Geert Litjens, Bjoern Menze, Olaf Ronneberger, Ronald~M Summers, et~al.
\newblock The medical segmentation decathlon.
\newblock \emph{Nature communications}, 13\penalty0 (1):\penalty0 4128, 2022.

\bibitem[Bankman(2008)]{bankman2008handbook}
Isaac Bankman.
\newblock \emph{Handbook of medical image processing and analysis}.
\newblock Elsevier, 2008.

\bibitem[Bilic et~al.(2019)Bilic, Christ, Vorontsov, Chlebus, Chen, Dou, Fu, Han, Heng, Hesser, et~al.]{bilic2019liver}
Patrick Bilic, Patrick~Ferdinand Christ, Eugene Vorontsov, Grzegorz Chlebus, Hao Chen, Qi Dou, Chi-Wing Fu, Xiao Han, Pheng-Ann Heng, J{\"u}rgen Hesser, et~al.
\newblock The liver tumor segmentation benchmark (lits).
\newblock \emph{arXiv preprint arXiv:1901.04056}, 2019.

\bibitem[Cabezas et~al.(2011)Cabezas, Oliver, Llad{\'o}, Freixenet, and Cuadra]{cabezas2011review}
Mariano Cabezas, Arnau Oliver, Xavier Llad{\'o}, Jordi Freixenet, and Meritxell~Bach Cuadra.
\newblock A review of atlas-based segmentation for magnetic resonance brain images.
\newblock \emph{Computer methods and programs in biomedicine}, 104\penalty0 (3):\penalty0 e158--e177, 2011.

\bibitem[Campello et~al.(2021)Campello, Gkontra, Izquierdo, Martin-Isla, Sojoudi, Full, Maier-Hein, Zhang, He, Ma, et~al.]{campello2021multi}
Victor~M Campello, Polyxeni Gkontra, Cristian Izquierdo, Carlos Martin-Isla, Alireza Sojoudi, Peter~M Full, Klaus Maier-Hein, Yao Zhang, Zhiqiang He, Jun Ma, et~al.
\newblock Multi-centre, multi-vendor and multi-disease cardiac segmentation: the m\&ms challenge.
\newblock \emph{IEEE Transactions on Medical Imaging}, 40\penalty0 (12):\penalty0 3543--3554, 2021.

\bibitem[Chang et~al.(2022)Chang, Yan, Zhou, Liu, Sawalha, Ye, Zhangli, Kanski, Al’Aref, Axel, et~al.]{chang2022deeprecon}
Qi Chang, Zhennan Yan, Mu Zhou, Di Liu, Khalid Sawalha, Meng Ye, Qilong Zhangli, Mikael Kanski, Subhi Al’Aref, Leon Axel, et~al.
\newblock Deeprecon: Joint 2d cardiac segmentation and 3d volume reconstruction via a structure-specific generative method.
\newblock In \emph{International Conference on Medical Image Computing and Computer-Assisted Intervention}, pages 567--577. Springer, 2022.

\bibitem[Chen et~al.(2018)Chen, Zhu, Papandreou, Schroff, and Adam]{chen2018encoder}
Liang-Chieh Chen, Yukun Zhu, George Papandreou, Florian Schroff, and Hartwig Adam.
\newblock Encoder-decoder with atrous separable convolution for semantic image segmentation.
\newblock In \emph{Proceedings of the European conference on computer vision (ECCV)}, pages 801--818, 2018.

\bibitem[Chen et~al.(2019{\natexlab{a}})Chen, Ma, and Zheng]{chen2019med3d}
Sihong Chen, Kai Ma, and Yefeng Zheng.
\newblock Med3d: Transfer learning for 3d medical image analysis.
\newblock \emph{arXiv preprint arXiv:1904.00625}, 2019{\natexlab{a}}.

\bibitem[Chen et~al.(2019{\natexlab{b}})Chen, Gao, Li, Zhao, and Zhao]{chen2019vertebrae}
Yizhi Chen, Yunhe Gao, Kang Li, Liang Zhao, and Jun Zhao.
\newblock Vertebrae identification and localization utilizing fully convolutional networks and a hidden markov model.
\newblock \emph{IEEE Transactions on Medical Imaging}, 39\penalty0 (2):\penalty0 387--399, 2019{\natexlab{b}}.

\bibitem[Cheng et~al.(2021)Cheng, Schwing, and Kirillov]{cheng2021per}
Bowen Cheng, Alex Schwing, and Alexander Kirillov.
\newblock Per-pixel classification is not all you need for semantic segmentation.
\newblock \emph{Advances in Neural Information Processing Systems}, 34:\penalty0 17864--17875, 2021.

\bibitem[Cheng et~al.(2022)Cheng, Misra, Schwing, Kirillov, and Girdhar]{cheng2022masked}
Bowen Cheng, Ishan Misra, Alexander~G Schwing, Alexander Kirillov, and Rohit Girdhar.
\newblock Masked-attention mask transformer for universal image segmentation.
\newblock In \emph{Proceedings of the IEEE/CVF Conference on Computer Vision and Pattern Recognition}, pages 1290--1299, 2022.

\bibitem[Cootes et~al.(1995)Cootes, Taylor, Cooper, and Graham]{cootes1995active}
Timothy~F Cootes, Christopher~J Taylor, David~H Cooper, and Jim Graham.
\newblock Active shape models-their training and application.
\newblock \emph{Computer vision and image understanding}, 61\penalty0 (1):\penalty0 38--59, 1995.

\bibitem[De~Fauw et~al.(2018)De~Fauw, Ledsam, Romera-Paredes, Nikolov, Tomasev, Blackwell, Askham, Glorot, O’Donoghue, Visentin, et~al.]{de2018clinically}
Jeffrey De~Fauw, Joseph~R Ledsam, Bernardino Romera-Paredes, Stanislav Nikolov, Nenad Tomasev, Sam Blackwell, Harry Askham, Xavier Glorot, Brendan O’Donoghue, Daniel Visentin, et~al.
\newblock Clinically applicable deep learning for diagnosis and referral in retinal disease.
\newblock \emph{Nature medicine}, 24\penalty0 (9):\penalty0 1342--1350, 2018.

\bibitem[Dmitriev and Kaufman(2019)]{dmitriev2019learning}
Konstantin Dmitriev and Arie~E Kaufman.
\newblock Learning multi-class segmentations from single-class datasets.
\newblock In \emph{Proceedings of the IEEE/CVF Conference on Computer Vision and Pattern Recognition}, pages 9501--9511, 2019.

\bibitem[Dosovitskiy et~al.(2020)Dosovitskiy, Beyer, Kolesnikov, Weissenborn, Zhai, Unterthiner, Dehghani, Minderer, Heigold, Gelly, et~al.]{dosovitskiy2020image}
Alexey Dosovitskiy, Lucas Beyer, Alexander Kolesnikov, Dirk Weissenborn, Xiaohua Zhai, Thomas Unterthiner, Mostafa Dehghani, Matthias Minderer, Georg Heigold, Sylvain Gelly, et~al.
\newblock An image is worth 16x16 words: Transformers for image recognition at scale.
\newblock \emph{arXiv preprint arXiv:2010.11929}, 2020.

\bibitem[Gao et~al.(2019{\natexlab{a}})Gao, Huang, Chen, Wang, Deng, Chen, Yang, Zhang, Tao, and Li]{gao2019focusnet}
Yunhe Gao, Rui Huang, Ming Chen, Zhe Wang, Jincheng Deng, Yuanyuan Chen, Yiwei Yang, Jie Zhang, Chanjuan Tao, and Hongsheng Li.
\newblock Focusnet: Imbalanced large and small organ segmentation with an end-to-end deep neural network for head and neck ct images.
\newblock In \emph{Medical Image Computing and Computer Assisted Intervention--MICCAI 2019: 22nd International Conference, Shenzhen, China, October 13--17, 2019, Proceedings, Part III 22}, pages 829--838. Springer, 2019{\natexlab{a}}.

\bibitem[Gao et~al.(2019{\natexlab{b}})Gao, Liu, and Zhao]{gao2019multi}
Yunhe Gao, Chang Liu, and Liang Zhao.
\newblock Multi-resolution path cnn with deep supervision for intervertebral disc localization and segmentation.
\newblock In \emph{Medical Image Computing and Computer Assisted Intervention--MICCAI 2019: 22nd International Conference, Shenzhen, China, October 13--17, 2019, Proceedings, Part II 22}, pages 309--317. Springer, 2019{\natexlab{b}}.

\bibitem[Gao et~al.(2021{\natexlab{a}})Gao, Huang, Yang, Zhang, Shao, Tao, Chen, Metaxas, Li, and Chen]{gao2021focusnetv2}
Yunhe Gao, Rui Huang, Yiwei Yang, Jie Zhang, Kainan Shao, Changjuan Tao, Yuanyuan Chen, Dimitris~N Metaxas, Hongsheng Li, and Ming Chen.
\newblock Focusnetv2: Imbalanced large and small organ segmentation with adversarial shape constraint for head and neck ct images.
\newblock \emph{Medical Image Analysis}, 67:\penalty0 101831, 2021{\natexlab{a}}.

\bibitem[Gao et~al.(2021{\natexlab{b}})Gao, Tang, Zhou, and Metaxas]{gao2021enabling}
Yunhe Gao, Zhiqiang Tang, Mu Zhou, and Dimitris Metaxas.
\newblock Enabling data diversity: efficient automatic augmentation via regularized adversarial training.
\newblock In \emph{International Conference on Information Processing in Medical Imaging}, pages 85--97. Springer, 2021{\natexlab{b}}.

\bibitem[Gao et~al.(2021{\natexlab{c}})Gao, Zhou, and Metaxas]{gao2021utnet}
Yunhe Gao, Mu Zhou, and Dimitris~N Metaxas.
\newblock Utnet: a hybrid transformer architecture for medical image segmentation.
\newblock In \emph{Medical Image Computing and Computer Assisted Intervention--MICCAI 2021: 24th International Conference, Strasbourg, France, September 27--October 1, 2021, Proceedings, Part III 24}, pages 61--71. Springer, 2021{\natexlab{c}}.

\bibitem[Gao et~al.(2022)Gao, Zhou, Liu, Yan, Zhang, and Metaxas]{gao2022data}
Yunhe Gao, Mu Zhou, Di Liu, Zhennan Yan, Shaoting Zhang, and Dimitris~N Metaxas.
\newblock A data-scalable transformer for medical image segmentation: architecture, model efficiency, and benchmark.
\newblock \emph{arXiv preprint arXiv:2203.00131}, 2022.

\bibitem[Gatidis et~al.(2022)Gatidis, Hepp, Fr{\"u}h, La~Foug{\`e}re, Nikolaou, Pfannenberg, Sch{\"o}lkopf, K{\"u}stner, Cyran, and Rubin]{gatidis2022whole}
Sergios Gatidis, Tobias Hepp, Marcel Fr{\"u}h, Christian La~Foug{\`e}re, Konstantin Nikolaou, Christina Pfannenberg, Bernhard Sch{\"o}lkopf, Thomas K{\"u}stner, Clemens Cyran, and Daniel Rubin.
\newblock A whole-body fdg-pet/ct dataset with manually annotated tumor lesions.
\newblock \emph{Scientific Data}, 9\penalty0 (1):\penalty0 601, 2022.

\bibitem[Hatamizadeh et~al.(2022)Hatamizadeh, Nath, Tang, Yang, Roth, and Xu]{hatamizadeh2022swin}
Ali Hatamizadeh, Vishwesh Nath, Yucheng Tang, Dong Yang, Holger~R Roth, and Daguang Xu.
\newblock Swin unetr: Swin transformers for semantic segmentation of brain tumors in mri images.
\newblock In \emph{Brainlesion: Glioma, Multiple Sclerosis, Stroke and Traumatic Brain Injuries: 7th International Workshop, BrainLes 2021, Held in Conjunction with MICCAI 2021, Virtual Event, September 27, 2021, Revised Selected Papers, Part I}, pages 272--284. Springer, 2022.

\bibitem[He et~al.(2016)He, Zhang, Ren, and Sun]{he2016deep}
Kaiming He, Xiangyu Zhang, Shaoqing Ren, and Jian Sun.
\newblock Deep residual learning for image recognition.
\newblock In \emph{Proceedings of the IEEE conference on computer vision and pattern recognition}, pages 770--778, 2016.

\bibitem[He et~al.(2022)He, Chen, Xie, Li, Doll{\'a}r, and Girshick]{he2022masked}
Kaiming He, Xinlei Chen, Saining Xie, Yanghao Li, Piotr Doll{\'a}r, and Ross Girshick.
\newblock Masked autoencoders are scalable vision learners.
\newblock In \emph{Proceedings of the IEEE/CVF Conference on Computer Vision and Pattern Recognition}, pages 16000--16009, 2022.

\bibitem[He et~al.(2023)He, Bao, Li, Grant, and Ou]{he2023accuracy}
Sheng He, Rina Bao, Jingpeng Li, P~Ellen Grant, and Yangming Ou.
\newblock Accuracy of segment-anything model (sam) in medical image segmentation tasks.
\newblock \emph{arXiv preprint arXiv:2304.09324}, 2023.

\bibitem[Heller et~al.(2019)Heller, Sathianathen, Kalapara, Walczak, Moore, Kaluzniak, Rosenberg, Blake, Rengel, Oestreich, et~al.]{heller2019kits19}
Nicholas Heller, Niranjan Sathianathen, Arveen Kalapara, Edward Walczak, Keenan Moore, Heather Kaluzniak, Joel Rosenberg, Paul Blake, Zachary Rengel, Makinna Oestreich, et~al.
\newblock The kits19 challenge data: 300 kidney tumor cases with clinical context, ct semantic segmentations, and surgical outcomes.
\newblock \emph{arXiv preprint arXiv:1904.00445}, 2019.

\bibitem[Huang et~al.(2020)Huang, Zheng, Hu, Zhang, and Li]{huang2020multi}
Rui Huang, Yuanjie Zheng, Zhiqiang Hu, Shaoting Zhang, and Hongsheng Li.
\newblock Multi-organ segmentation via co-training weight-averaged models from few-organ datasets.
\newblock In \emph{Medical Image Computing and Computer Assisted Intervention--MICCAI 2020: 23rd International Conference, Lima, Peru, October 4--8, 2020, Proceedings, Part IV 23}, pages 146--155. Springer, 2020.

\bibitem[Hussain et~al.(2022)Hussain, Mubeen, Ullah, Shah, Khan, Zahoor, Ullah, Khan, and Sultan]{hussain2022modern}
Shah Hussain, Iqra Mubeen, Niamat Ullah, Syed Shahab Ud~Din Shah, Bakhtawar~Abduljalil Khan, Muhammad Zahoor, Riaz Ullah, Farhat~Ali Khan, and Mujeeb~A Sultan.
\newblock Modern diagnostic imaging technique applications and risk factors in the medical field: A review.
\newblock \emph{BioMed Research International}, 2022, 2022.

\bibitem[Iglesias and Sabuncu(2015)]{iglesias2015multi}
Juan~Eugenio Iglesias and Mert~R Sabuncu.
\newblock Multi-atlas segmentation of biomedical images: a survey.
\newblock \emph{Medical image analysis}, 24\penalty0 (1):\penalty0 205--219, 2015.

\bibitem[Isensee et~al.(2021)Isensee, Jaeger, Kohl, Petersen, and Maier-Hein]{isensee2021nnu}
Fabian Isensee, Paul~F Jaeger, Simon~AA Kohl, Jens Petersen, and Klaus~H Maier-Hein.
\newblock nnu-net: a self-configuring method for deep learning-based biomedical image segmentation.
\newblock \emph{Nature methods}, 18\penalty0 (2):\penalty0 203--211, 2021.

\bibitem[Ji et~al.(2022)Ji, Bai, Yang, Ge, Zhu, Zhang, Li, Zhang, Ma, Wan, et~al.]{ji2022amos}
Yuanfeng Ji, Haotian Bai, Jie Yang, Chongjian Ge, Ye Zhu, Ruimao Zhang, Zhen Li, Lingyan Zhang, Wanling Ma, Xiang Wan, et~al.
\newblock Amos: A large-scale abdominal multi-organ benchmark for versatile medical image segmentation.
\newblock \emph{arXiv preprint arXiv:2206.08023}, 2022.

\bibitem[Kamnitsas et~al.(2017)Kamnitsas, Ledig, Newcombe, Simpson, Kane, Menon, Rueckert, and Glocker]{kamnitsas2017efficient}
Konstantinos Kamnitsas, Christian Ledig, Virginia~FJ Newcombe, Joanna~P Simpson, Andrew~D Kane, David~K Menon, Daniel Rueckert, and Ben Glocker.
\newblock Efficient multi-scale 3d cnn with fully connected crf for accurate brain lesion segmentation.
\newblock \emph{Medical image analysis}, 36:\penalty0 61--78, 2017.

\bibitem[Kaplan et~al.(2020)Kaplan, McCandlish, Henighan, Brown, Chess, Child, Gray, Radford, Wu, and Amodei]{kaplan2020scaling}
Jared Kaplan, Sam McCandlish, Tom Henighan, Tom~B Brown, Benjamin Chess, Rewon Child, Scott Gray, Alec Radford, Jeffrey Wu, and Dario Amodei.
\newblock Scaling laws for neural language models.
\newblock \emph{arXiv preprint arXiv:2001.08361}, 2020.

\bibitem[Kavur et~al.(2021)Kavur, Gezer, Barış, Aslan, Conze, Groza, Pham, Chatterjee, Ernst, Özkan, Baydar, Lachinov, Han, Pauli, Isensee, Perkonigg, Sathish, Rajan, Sheet, Dovletov, Speck, Nürnberger, Maier-Hein, {Bozdağı Akar}, Ünal, Dicle, and Selver]{CHAOS2021}
A.~Emre Kavur, N.~Sinem Gezer, Mustafa Barış, Sinem Aslan, Pierre-Henri Conze, Vladimir Groza, Duc~Duy Pham, Soumick Chatterjee, Philipp Ernst, Savaş Özkan, Bora Baydar, Dmitry Lachinov, Shuo Han, Josef Pauli, Fabian Isensee, Matthias Perkonigg, Rachana Sathish, Ronnie Rajan, Debdoot Sheet, Gurbandurdy Dovletov, Oliver Speck, Andreas Nürnberger, Klaus~H. Maier-Hein, Gözde {Bozdağı Akar}, Gözde Ünal, Oğuz Dicle, and M.~Alper Selver.
\newblock {CHAOS Challenge - combined (CT-MR) healthy abdominal organ segmentation}.
\newblock \emph{Medical Image Analysis}, 69:\penalty0 101950, 2021.

\bibitem[Kirillov et~al.(2023)Kirillov, Mintun, Ravi, Mao, Rolland, Gustafson, Xiao, Whitehead, Berg, Lo, et~al.]{kirillov2023segment}
Alexander Kirillov, Eric Mintun, Nikhila Ravi, Hanzi Mao, Chloe Rolland, Laura Gustafson, Tete Xiao, Spencer Whitehead, Alexander~C Berg, Wan-Yen Lo, et~al.
\newblock Segment anything.
\newblock \emph{arXiv preprint arXiv:2304.02643}, 2023.

\bibitem[Lambert et~al.(2020)Lambert, Petitjean, Dubray, and Kuan]{lambert2020segthor}
Zo{\'e} Lambert, Caroline Petitjean, Bernard Dubray, and Su Kuan.
\newblock Segthor: Segmentation of thoracic organs at risk in ct images.
\newblock In \emph{2020 Tenth International Conference on Image Processing Theory, Tools and Applications (IPTA)}, pages 1--6. IEEE, 2020.

\bibitem[Landman et~al.()Landman, Xu, Lgelsias, Styner, Langerak, and Arno]{bcv}
Bennett Landman, Zhoubing Xu, Juan Lgelsias, Martin Styner, Thomas Langerak, and Klein Arno.
\newblock Multi-atlas labeling beyond the cranial vault - workshop and challenge.

\bibitem[Li et~al.()Li, Zhou, Deng, and Chen]{structseg}
Hongsheng Li, Jinghao Zhou, Jincheng Deng, and Ming Chen.
\newblock Automatic structure segmentation for radiotherapy planning challenge 2019.

\bibitem[Lin et~al.(2017)Lin, Doll{\'a}r, Girshick, He, Hariharan, and Belongie]{lin2017feature}
Tsung-Yi Lin, Piotr Doll{\'a}r, Ross Girshick, Kaiming He, Bharath Hariharan, and Serge Belongie.
\newblock Feature pyramid networks for object detection.
\newblock In \emph{Proceedings of the IEEE conference on computer vision and pattern recognition}, pages 2117--2125, 2017.

\bibitem[Liu et~al.(2022)Liu, Gao, Zhangli, Han, He, Xia, Wen, Chang, Yan, Zhou, et~al.]{liu2022transfusion}
Di Liu, Yunhe Gao, Qilong Zhangli, Ligong Han, Xiaoxiao He, Zhaoyang Xia, Song Wen, Qi Chang, Zhennan Yan, Mu Zhou, et~al.
\newblock Transfusion: multi-view divergent fusion for medical image segmentation with transformers.
\newblock In \emph{International Conference on Medical Image Computing and Computer-Assisted Intervention}, pages 485--495. Springer, 2022.

\bibitem[Liu et~al.(2023)Liu, Zhang, Chen, Xiao, Lu, Landman, Yuan, Yuille, Tang, and Zhou]{liu2023clip}
Jie Liu, Yixiao Zhang, Jie-Neng Chen, Junfei Xiao, Yongyi Lu, Bennett~A Landman, Yixuan Yuan, Alan Yuille, Yucheng Tang, and Zongwei Zhou.
\newblock Clip-driven universal model for organ segmentation and tumor detection.
\newblock \emph{arXiv preprint arXiv:2301.00785}, 2023.

\bibitem[Liu et~al.(2021)Liu, Lin, Cao, Hu, Wei, Zhang, Lin, and Guo]{liu2021swin}
Ze Liu, Yutong Lin, Yue Cao, Han Hu, Yixuan Wei, Zheng Zhang, Stephen Lin, and Baining Guo.
\newblock Swin transformer: Hierarchical vision transformer using shifted windows.
\newblock In \emph{Proceedings of the IEEE/CVF international conference on computer vision}, pages 10012--10022, 2021.

\bibitem[Ma and Wang(2023)]{ma2023segment}
Jun Ma and Bo Wang.
\newblock Segment anything in medical images.
\newblock \emph{arXiv preprint arXiv:2304.12306}, 2023.

\bibitem[Nestle et~al.(2005)Nestle, Kremp, Schaefer-Schuler, Sebastian-Welsch, Hellwig, R{\"u}be, and Kirsch]{nestle2005comparison}
Ursula Nestle, Stephanie Kremp, Andrea Schaefer-Schuler, Christiane Sebastian-Welsch, Dirk Hellwig, Christian R{\"u}be, and Carl-Martin Kirsch.
\newblock Comparison of different methods for delineation of 18f-fdg pet--positive tissue for target volume definition in radiotherapy of patients with non--small cell lung cancer.
\newblock \emph{Journal of nuclear medicine}, 46\penalty0 (8):\penalty0 1342--1348, 2005.

\bibitem[Oktay et~al.(2017)Oktay, Ferrante, Kamnitsas, Heinrich, Bai, Caballero, Cook, De~Marvao, Dawes, O‘Regan, et~al.]{oktay2017anatomically}
Ozan Oktay, Enzo Ferrante, Konstantinos Kamnitsas, Mattias Heinrich, Wenjia Bai, Jose Caballero, Stuart~A Cook, Antonio De~Marvao, Timothy Dawes, Declan~P O‘Regan, et~al.
\newblock Anatomically constrained neural networks (acnns): application to cardiac image enhancement and segmentation.
\newblock \emph{IEEE transactions on medical imaging}, 37\penalty0 (2):\penalty0 384--395, 2017.

\bibitem[Raghu et~al.(2019)Raghu, Zhang, Kleinberg, and Bengio]{raghu2019transfusion}
Maithra Raghu, Chiyuan Zhang, Jon Kleinberg, and Samy Bengio.
\newblock Transfusion: Understanding transfer learning for medical imaging.
\newblock \emph{Advances in neural information processing systems}, 32, 2019.

\bibitem[Rao et~al.(2022)Rao, Wan, Arabshahi, Ma, Lee, Tian, Zhang, Laine, and Guo]{rao2022improving}
Vishwanatha~M Rao, Zihan Wan, Soroush Arabshahi, David~J Ma, Pin-Yu Lee, Ye Tian, Xuzhe Zhang, Andrew~F Laine, and Jia Guo.
\newblock Improving across-dataset brain tissue segmentation for mri imaging using transformer.
\newblock \emph{Frontiers in Neuroimaging}, 1:\penalty0 1023481, 2022.

\bibitem[Razzak et~al.(2018)Razzak, Naz, and Zaib]{razzak2018deep}
Muhammad~Imran Razzak, Saeeda Naz, and Ahmad Zaib.
\newblock Deep learning for medical image processing: Overview, challenges and the future.
\newblock \emph{Classification in BioApps: Automation of Decision Making}, pages 323--350, 2018.

\bibitem[Rodrigue et~al.(2012)Rodrigue, Kennedy, Devous, Rieck, Hebrank, Diaz-Arrastia, Mathews, and Park]{rodrigue2012beta}
KM Rodrigue, KM Kennedy, MD Devous, JR Rieck, AC Hebrank, R Diaz-Arrastia, D Mathews, and DC Park.
\newblock $\beta$-amyloid burden in healthy aging: regional distribution and cognitive consequences.
\newblock \emph{Neurology}, 78\penalty0 (6):\penalty0 387--395, 2012.

\bibitem[Ronneberger et~al.(2015)Ronneberger, Fischer, and Brox]{ronneberger2015u}
Olaf Ronneberger, Philipp Fischer, and Thomas Brox.
\newblock U-net: Convolutional networks for biomedical image segmentation.
\newblock In \emph{Medical Image Computing and Computer-Assisted Intervention--MICCAI 2015: 18th International Conference, Munich, Germany, October 5-9, 2015, Proceedings, Part III 18}, pages 234--241. Springer, 2015.

\bibitem[Sarkany et~al.(2018)Sarkany, Shenoy-Bhangle, Catanzano, Fineberg, Eisenberg, and Slanetz]{sarkany2018running}
David~S Sarkany, Anuradha~S Shenoy-Bhangle, Tara~M Catanzano, Tabitha~A Fineberg, Ronald~L Eisenberg, and Priscilla~J Slanetz.
\newblock Running a radiology residency program: strategies for success.
\newblock \emph{Radiographics}, 38\penalty0 (6):\penalty0 1729--1743, 2018.

\bibitem[Scanlon and Sanders(2018)]{scanlon2018essentials}
Valerie~C Scanlon and Tina Sanders.
\newblock \emph{Essentials of anatomy and physiology}.
\newblock FA Davis, 2018.

\bibitem[Shen et~al.(2015)Shen, Zhou, Yang, Yang, and Tian]{shen2015multi}
Wei Shen, Mu Zhou, Feng Yang, Caiyun Yang, and Jie Tian.
\newblock Multi-scale convolutional neural networks for lung nodule classification.
\newblock In \emph{Information Processing in Medical Imaging: 24th International Conference, IPMI 2015, Sabhal Mor Ostaig, Isle of Skye, UK, June 28-July 3, 2015, Proceedings 24}, pages 588--599. Springer, 2015.

\bibitem[Sohn et~al.(2020)Sohn, Berthelot, Carlini, Zhang, Zhang, Raffel, Cubuk, Kurakin, and Li]{sohn2020fixmatch}
Kihyuk Sohn, David Berthelot, Nicholas Carlini, Zizhao Zhang, Han Zhang, Colin~A Raffel, Ekin~Dogus Cubuk, Alexey Kurakin, and Chun-Liang Li.
\newblock Fixmatch: Simplifying semi-supervised learning with consistency and confidence.
\newblock \emph{Advances in neural information processing systems}, 33:\penalty0 596--608, 2020.

\bibitem[Tarvainen and Valpola(2017)]{tarvainen2017mean}
Antti Tarvainen and Harri Valpola.
\newblock Mean teachers are better role models: Weight-averaged consistency targets improve semi-supervised deep learning results.
\newblock \emph{Advances in neural information processing systems}, 30, 2017.

\bibitem[Ulrich et~al.(2023)Ulrich, Isensee, Wald, Zenk, Baumgartner, and Maier-Hein]{ulrich2023multitalent}
Constantin Ulrich, Fabian Isensee, Tassilo Wald, Maximilian Zenk, Michael Baumgartner, and Klaus~H Maier-Hein.
\newblock Multitalent: A multi-dataset approach to medical image segmentation.
\newblock In \emph{International Conference on Medical Image Computing and Computer-Assisted Intervention}, pages 648--658. Springer, 2023.

\bibitem[Xie et~al.(2022)Xie, Zhang, Xia, and Wu]{xie2022unimiss}
Yutong Xie, Jianpeng Zhang, Yong Xia, and Qi Wu.
\newblock Unimiss: Universal medical self-supervised learning via breaking dimensionality barrier.
\newblock In \emph{European Conference on Computer Vision}, pages 558--575. Springer, 2022.

\bibitem[Ye et~al.(2022)Ye, Zhang, Chen, and Xia]{ye2022desd}
Yiwen Ye, Jianpeng Zhang, Ziyang Chen, and Yong Xia.
\newblock Desd: Self-supervised learning with deep self-distillation for 3d medical image segmentation.
\newblock In \emph{International Conference on Medical Image Computing and Computer-Assisted Intervention}, pages 545--555. Springer, 2022.

\bibitem[Ye et~al.(2023)Ye, Xie, Zhang, Chen, and Xia]{ye2023uniseg}
Yiwen Ye, Yutong Xie, Jianpeng Zhang, Ziyang Chen, and Yong Xia.
\newblock Uniseg: A prompt-driven universal segmentation model as well as a strong representation learner.
\newblock \emph{arXiv preprint arXiv:2304.03493}, 2023.

\bibitem[You et~al.(2019)You, Li, Reddi, Hseu, Kumar, Bhojanapalli, Song, Demmel, Keutzer, and Hsieh]{you2019large}
Yang You, Jing Li, Sashank Reddi, Jonathan Hseu, Sanjiv Kumar, Srinadh Bhojanapalli, Xiaodan Song, James Demmel, Kurt Keutzer, and Cho-Jui Hsieh.
\newblock Large batch optimization for deep learning: Training bert in 76 minutes.
\newblock \emph{arXiv preprint arXiv:1904.00962}, 2019.

\bibitem[Zhang et~al.(2021{\natexlab{a}})Zhang, Xie, Xia, and Shen]{zhang2021dodnet}
Jianpeng Zhang, Yutong Xie, Yong Xia, and Chunhua Shen.
\newblock Dodnet: Learning to segment multi-organ and tumors from multiple partially labeled datasets.
\newblock In \emph{Proceedings of the IEEE/CVF conference on computer vision and pattern recognition}, pages 1195--1204, 2021{\natexlab{a}}.

\bibitem[Zhang et~al.(2021{\natexlab{b}})Zhang, Pang, Chen, and Loy]{zhang2021k}
Wenwei Zhang, Jiangmiao Pang, Kai Chen, and Chen~Change Loy.
\newblock K-net: Towards unified image segmentation.
\newblock \emph{Advances in Neural Information Processing Systems}, 34:\penalty0 10326--10338, 2021{\natexlab{b}}.

\bibitem[Zhangli et~al.(2022)Zhangli, Yi, Liu, He, Xia, Chang, Han, Gao, Wen, Tang, et~al.]{zhangli2022region}
Qilong Zhangli, Jingru Yi, Di Liu, Xiaoxiao He, Zhaoyang Xia, Qi Chang, Ligong Han, Yunhe Gao, Song Wen, Haiming Tang, et~al.
\newblock Region proposal rectification towards robust instance segmentation of biological images.
\newblock In \emph{International Conference on Medical Image Computing and Computer-Assisted Intervention}, pages 129--139. Springer, 2022.

\bibitem[Zhou et~al.(2019)Zhou, Li, Bai, Wang, Chen, Han, Fishman, and Yuille]{zhou2019prior}
Yuyin Zhou, Zhe Li, Song Bai, Chong Wang, Xinlei Chen, Mei Han, Elliot Fishman, and Alan~L Yuille.
\newblock Prior-aware neural network for partially-supervised multi-organ segmentation.
\newblock In \emph{Proceedings of the IEEE/CVF international conference on computer vision}, pages 10672--10681, 2019.

\end{thebibliography}
}

\clearpage
\setcounter{page}{1}
\maketitlesupplementary

\section{Supplementary}
\label{sec:supp}
\begin{figure*}[h]
    \centering
    \includegraphics[width=1\textwidth]{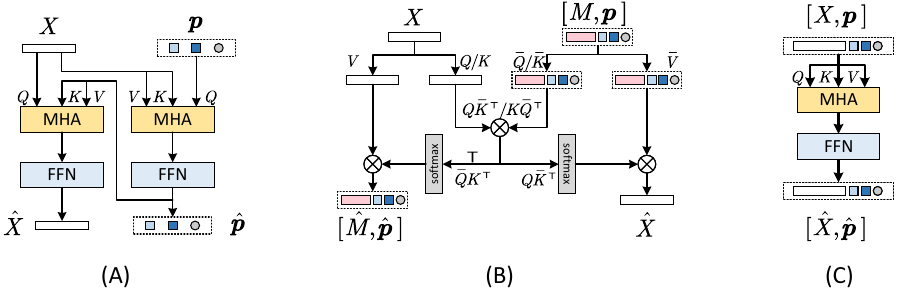}
    \caption{The implementation of prior fusion module for existing backbones. (A) The implementation for CNN backbones, like ResUNet~\cite{ronneberger2015u}. We use a bi-directional cross-attention module to process both the feature map $X$ and the prior tokens $\boldsymbol{p}$. (B) The implementation for MedFormer~\cite{gao2022data}. We merge the prior tokens into the semantic map of its B-MHA module. (C) The implementation for conventional attention module, e.g. ViT~\cite{dosovitskiy2020image} backbone. The normalization layers and residue connections for all three implementations are omitted for simplicity.}
    \label{fig:appendix_priorfusion}
\end{figure*}

\subsection{Prior fusion module details} The prior fusion module is an essential part of Hermes, which is responsible for integrating knowledge encapsulated in prior tokens with image feature maps. This fusion process utilizes attention modules, enabling efficient and global information exchange between prior tokens and feature maps. This design ensures Hermes' compatibility with existing backbones. We illustrate several implementations for common backbones in Figure \ref{fig:appendix_priorfusion}. We first introduce the formulation with the conventional attention module. Then, we introduce the bi-directional cross-attention for CNN backbones. Finally, we show how to merge the prior fusion into the MedFormer backbone.

\textbf{Conventional attention module.} Conventional Transformer architectures (e.g. ViT~\cite{dosovitskiy2020image}) often rely on multi-head self-attention that captures the all-to-all pairwise dependencies among input tokens. To extend, we present a general formulation of merging prior fusion into a conventional attention module, see Figure \ref{fig:appendix_priorfusion} (C). Given the input feature map $X\in\mathcal{R}^{n\times C}$ ($n=D\times H\times W$ is the number of tokens in the feature map, $C$ denotes the token dimension), and the prior tokens $\boldsymbol{p}\in\mathcal{R}^{(|\boldsymbol{t}_{k}|+l)\times C}$, where $|\boldsymbol{t}_{k}|$ denotes the number of tasks, and $l$ is the length of each modality prior. We first concatenate the feature map and the prior tokens together as the input of the transformer block: $I=[X, \boldsymbol{p}]\in\mathcal{R}^{n+|\boldsymbol{t}_{k}|+l}$.  The conventional Transformer block operates as follows:

\begin{equation}
    \begin{aligned}
        &I' = \text{LN}(I)\\
        &I'' = \text{MHA}(I', I', I')) + I \\
        &\text{MHA}(q, k, v) = [head_1, \cdots, head_h]W^O \\
        &head_i = \text{Attention}(qW^Q_i, kW^K_i, vW^V_i) \\
        &\text{Attention}(\mathbf{Q}, \mathbf{K}, \mathbf{V}) = \text{softmax}(\frac{\mathbf{QK}^{\mathsf{T}}}{\sqrt{d_h}})\mathbf{V}\\
        &\hat{I} = \text{FFN}(\text{LN}(I'')) + I''      
    \end{aligned}
\end{equation}
where $W_Q, W^K, W^V, W^O$ are weight matrices, and $d_h$ is the dimension of each head. We can then obtain the prior-injected feature map and the posterior tokens by splitting $\hat{I}=[\hat{X}, \hat{\boldsymbol{p}}]$. The computation complexity for the above prior-modified attention module is $O((n+|\boldsymbol{t}_{k}|+l)^2)$

\textbf{CNN backbones.} CNN backbones, like ResUNet~\cite{ronneberger2015u,he2016deep}, usually use hierarchical architectures. The quadratic complexity of the above all-to-all attention makes it unaffordable to apply the prior fusion module on high-resolution feature maps (e.g. $n=32,768$ for a $32\times32\times32$ feature map). Therefore, we implement the prior fusion module with a bi-directional cross-attention module instead, see Figure \ref{fig:appendix_priorfusion} (A). The prior tokens $\boldsymbol{p}$ first aggregate image-specific information from the feature map $X$ to obtain the posterior tokens $\hat{\boldsymbol{p}}$:

\begin{equation}
    \begin{aligned}
        &\boldsymbol{p}' = \text{LN}(\boldsymbol{p})\\
        &X' = \text{LN}(X)\\
        &\boldsymbol{p}'' = \text{MHA}(\boldsymbol{p}', X', X') + \boldsymbol{p} \\
        &\hat{\boldsymbol{p}} = \text{FFN}(\text{LN}(\boldsymbol{p}'')) + \boldsymbol{p}''       
    \end{aligned}
\end{equation}

Then we inject the knowledge in the posterior tokens to obtain the prior-injected feature map $\hat{X}$:

\begin{equation}
    \begin{aligned}
        &\hat{\boldsymbol{p}}' = \text{LN}( \hat{\boldsymbol{p}})\\
        &X'' = \text{MHA}(X', \hat{\boldsymbol{p}}', \hat{\boldsymbol{p}}') + X \\
        & \hat{X} = \text{FFN}(\text{LN}(X'')) + X''
    \end{aligned}
\end{equation}

As $|\boldsymbol{t}_{k}|+l \ll n$, the computation complexity of the bi-direction cross-attention for the prior fusion module is $O(n)$. For example, for the dataset with the most tasks, AMOS CT, $|\boldsymbol{t}_{k}|+l=25 \ll n=32,768$ for a $32\times32\times32$ feature map. With this design, the prior fusion module can adaptively integrate the prior tokens and the feature maps with minor additional computational costs. We implement Hermes-R by inserting the cross-attention module at the end of each stage of the ResUNet, i.e. after the convolution layers at $4\times$, $8\times$, and $16\times$ downsampling scales.

\textbf{MedFormer.} MedFormer~\cite{gao2022data} is a Transformer model proposed for medical image segmentation. One key component of MedFormer is its B-MHA attention module, which incorporates a compressed semantic map to reduce computation complexity as well as enhance representation learning. In Figure \ref{fig:appendix_priorfusion} (B), we show our implementation of merging the prior fusion module into the B-MHA module of the MedFormer backbone. In B-MHA, $M$ is a semantic map that is encoded and refined for rich semantic information within a much lower spatial resolution compared to the feature map $X$. We concatenate the semantic map $M$ with the prior tokens $\boldsymbol{p}$: $I_M=[M, \boldsymbol{p}]$. The $X$ and  $I_M$ are linearly projected to $\mathbf{Q/K/V}$ and $\mathbf{\bar{Q}/\bar{K}/\bar{V}}$ respectively. To reduce the computation, B-MHA shares the query and key of $X$ and $I_M$, i.e. $\mathbf{Q}=\mathbf{K}$ and $\mathbf{\bar{Q}}=\mathbf{\bar{K}}$, as the dot product of the query and key in cross-attention measures the similarity of token pairs of two inputs, which is symmetrical. The attention matrix is reused by simply transposing the dot product matrix:

\begin{equation}
    \begin{aligned}
        &\hat{X} = \text{Attention}(\mathbf{Q}, \mathbf{\bar{K}}, \mathbf{\bar{V}})=\text{softmax}(\frac{\mathbf{Q}\mathbf{\bar{K}}^{\mathsf{T}}}{\sqrt{d}})\mathbf{\bar{V}}\\
        &\hat{I}_M = \text{Attention}(\mathbf{\bar{Q}}, \mathbf{K}, \mathbf{V})=\text{softmax}(\frac{\mathbf{\bar{Q}}\mathbf{K}^{\mathsf{T}}}{\sqrt{d}})\mathbf{V}\\
        &(\mathbf{Q}\mathbf{\bar{K}}^{\mathsf{T}})^{\mathsf{T}}=(\mathbf{K}\mathbf{\bar{Q}}^{\mathsf{T}})^{\mathsf{T}} = \mathbf{\bar{Q}}\mathbf{K}^{\mathsf{T}} 
    \end{aligned}
\end{equation}

The normalization layer, FFN, and residue connections are not included in the equation for simplicity. More details can be found in the original MedFormer paper. We can then obtain the updated semantic map and the posterior tokens by splitting $\hat{I}_M=[\hat{M}, \hat{\boldsymbol{p}}]$. As $|M|+|\boldsymbol{t}_k|+l\ll n$, the computation complexity is $O(n)$. We implement Hermes-M by incorporating the prior fusion module with the B-MHA module on the $4\times$, $8\times$ and $16\times$ downsampling scales within the MedFormer backbone.

\textbf{Computation comparison.} In Table \ref{tab:appendix_consumption}, we present the GPU memory usage, number of parameters, and inference time for Hermes with various backbones. For the ResUNet backbone, thanks to the efficient bi-directional cross-attention, Hermes-R only slightly increases GPU memory usage and inference time, despite additional parameters due to the cross-attention in the prior fusion module. For the Transformer backbone MedFormer, Hermes-M demonstrates almost identical consumption on memory, inference time, and the number of parameters compared with MedFormer. These results exemplify the efficacy of integrating the prior fusion module into MedFormer's existing attention module, highlighting Hermes' ability in leveraging different backbones without significantly impacting the required computational resources.

\begin{table}[h]
    \centering
    \small
    \caption{Computation comparison between the Hermes and the corresponding backbone. The memory consumption and inference time are measured with an image size of $2\times1\times128\times128\times128$ on one Nvidia A100 GPU. We report the average inference time over 100 runs.}\label{tab:appendix_consumption}
    \begin{tabular}{c|c|c|c}
    \toprule
    Model      & Memory/G & \#Params/M      & Inference Time/s  \\\midrule
    ResUNet    & 11.23    & 40.56           &  0.13 \\
    Hermes-R   & 11.54    & 59.61           & 0.16 \\
    MedFormer  & 11.44    & 43.20           & 0.19 \\
    Hermes-M   & 11.62    & 44.50           & 0.20 \\\bottomrule
    \end{tabular}

    \label{tab:my_label}
\end{table}

\section{Supplement Experiments}

\begin{figure*}[h]
    \centering
    \includegraphics[width=1\textwidth]{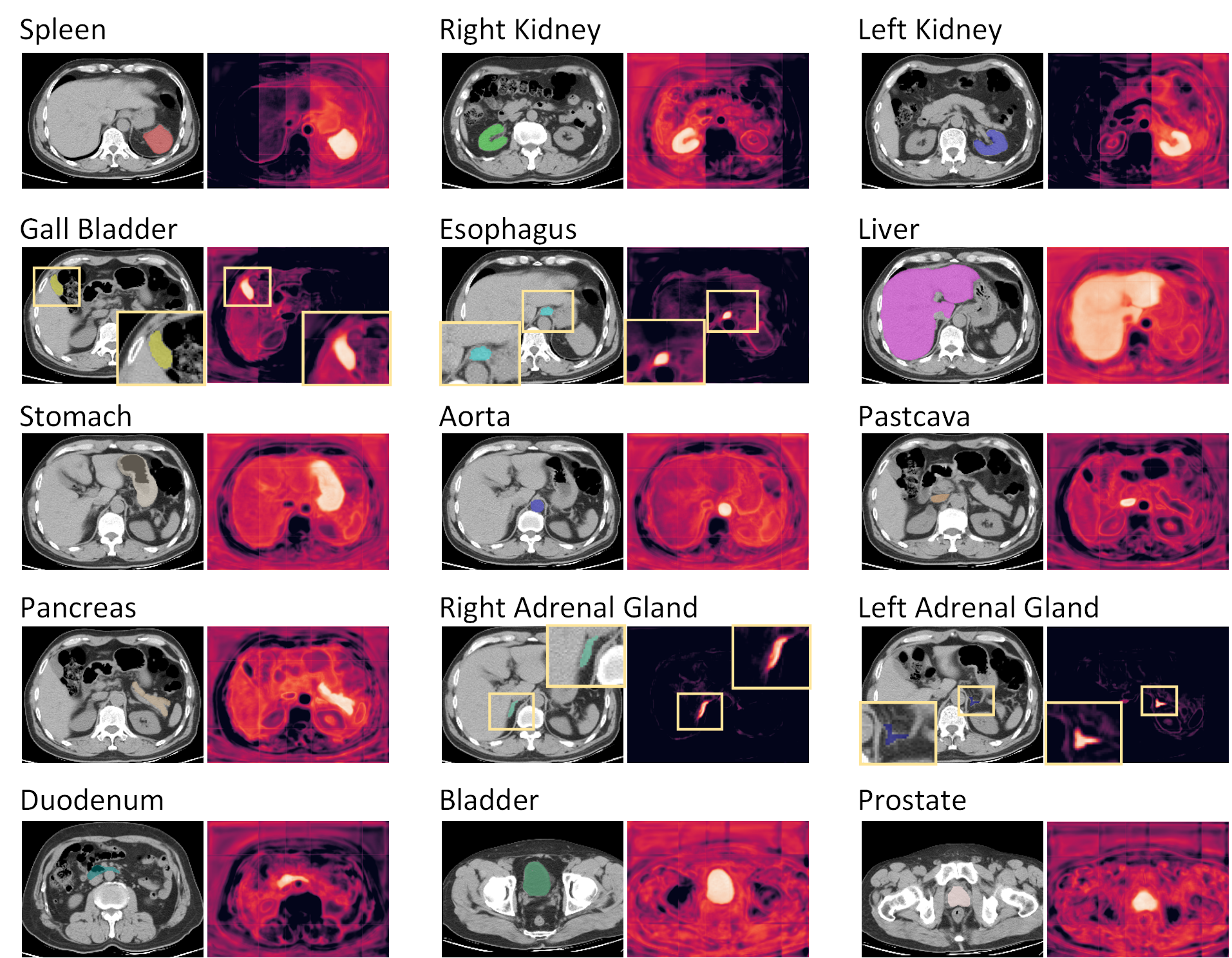}
    \caption{Heatmap visualization of each posterior prototype for the AMOS CT dataset. The brighter means the higher similarity. The yellow boxes are the zoomed-in version of small organs for better visualization.}
    \label{fig:appendix_visualization}
\end{figure*}

\textbf{Visual analysis on posterior prototypes.} In Figure \ref{fig:appendix_visualization}, we show the heatmap visualization of each posterior prototype with the output feature map of the decoder measured with dot-product similarity. The visualization underscores the quality of Hermes's predicted posterior prototypes, which adeptly capture the semantic essence of each category, aligning closely with the respective feature maps. Remarkably, Hermes manages precise predictions even for small organs with complex shapes. Take the right and left adrenal glands as an example: despite their tiny size and irregular shape, the posterior prototypes predicted by Hermes accurately reflect their intricate edges.

\textbf{Comparison with other methods.} We provide the detailed performance of other comparison methods on each dataset, see Table \ref{tab:appendix_comparison}. The ResUNet is trained under the traditional paradigm. All comparison methods are trained with the universal paradigm and are implemented with ResUNet for a fair comparison. All methods under the universal paradigm exhibit better performance compared with traditionally trained ResUNet. Among universal method settings, our Hermes-R shows a consistent advantage on the eleven upstream datasets.

The detailed number of Figure 3 (A) in the main text is presented in Table \ref{tab:appendix_number}. All methods under the universal paradigm show better than the ResUNet trained on each individual dataset. Hermes demonstrates consistent advantages, especially on difficult classes, modality PET, tumor\&lesion classes, and head\&neck region tasks.

\begin{table*}
  \caption{Comparison with other methods. ResUNet is trained with the traditional paradigm, while all comparison methods are reimplemented with the ResUNet backbone for fair comparison and extend to the universal medical image segmentation paradigm.}
  \label{tab:appendix_comparison}
  \centering
  \footnotesize
  \setlength{\tabcolsep}{1.8mm}{
  \begin{tabular}{c|cccccccccccc}
    \toprule
    Model    & BCV    & SS T & SS H & LiTS T    & KiTS T    & AMOS CT   & AMOS MR   & CHAOS & M\&Ms & AutoPET & DLBS & AVG       \\
    \hline
    
    ResUNet & 84.36	&88.59	& 78.12 &64.87	&81.89	&88.97	&85.43	&91.34 & 85.73 & 65.52 & 94.31 & 82.65  \\\hline
    Multi-decoder~\cite{chen2019med3d} & 83.90	&89.18	& 78.31 &65.74	&81.66	&89.27	&85.65	&91.56	&86.00 & 66.06 & 94.71 & 82.91 \\ 
    DoDNet~\cite{zhang2021dodnet} &85.02	&88.87 & 78.49	&65.84	&82.65	&88.86	&86.22	&91.35 & 85.97 &67.49 & 94.94	&83.25\\
    CLIP-driven~\cite{liu2023clip} &85.12	&89.34	& 78.50 &65.37	&82.83	&88.94	&86.39	&91.81 & 86.04 & 66.78 & 95.17	&83.30 \\
    UniSeg~\cite{ye2023uniseg} & 85.32 & 89.39 & 78.69 & 65.80 & 82.96 & 89.17 & 86.55 & 91.85 & 86.26 & 70.12 & 95.34 & 83.77 \\
    MultiTalent~\cite{ulrich2023multitalent} &  85.18 & 89.18 & 80.01 & 65.33 & 82.25 & 89.13 & 86.57 & 91.55 & 86.28 & 71.51 & 95.75 & 83.88\\
    \rowcolor{lightgray}
    Hermes-R & \bf 85.99	&\bf89.50 & \bf 80.62	&\bf 67.49	&\bf 85.53	&\bf 89.63	&\bf 86.78	&\bf 92.01 & \bf 86.94 &\bf 73.69 & \bf 96.21 &\bf 84.95\\

    \bottomrule
  \end{tabular}}
\end{table*}

\begin{table*}
  \caption{The detailed number of Figure 3 (A) in the main text. We compare traditionally trained ResUNet and other SOTA method under the universal paradigm in six aspects. "Difficult Classes" are the classes that have Dice scores lower than 80 under the traditional paradigm.}
  \label{tab:appendix_number}
  \centering
  \small
  \setlength{\tabcolsep}{2mm}{
  \begin{tabular}{c|cccccc}
    \toprule
    Model    & Difficult Classes & Modality PET & Tumor\&Lesion & Region Head\&Neck & All Classes & All Datasets     \\
    \hline
    ResUNet & 70.67 & 65.52 & 70.76 & 78.13 & 84.54 & 82.65 \\\hline
    Multi-decoder  & 71.42 & 66.06 & 71.15 & 78.31 & 84.70 & 82.91\\
    DoDNet & 72.17 & 67.49 & 71.99 & 78.48 & 84.93 & 83.25\\
    CLIP-driven & 72.11 & 66.78 & 71.64 & 78.50 & 85.05 & 83.30\\
    UniSeg  & 72.46 & 70.12 & 72.96 & 78.69 & 85.27 & 83.77\\
    MultiTalent & 73.18 & 71.51 & 73.03 & 80.01 & 85.57 & 83.88\\
    \rowcolor{lightgray}
    Hermes-R & 74.35 & 73.69 & 75.57 & 80.62 & 86.16 & 84.95\\

    \bottomrule
  \end{tabular}}
\end{table*}

\textbf{Ablation on the length of modality prior token.} We present an additional ablation study on the length of the modality prior token, see Table \ref{tab:appendix_modality_length}. We follow the setting in the ablation study section in the manuscript using the ResUNet-Small backbone to implement Hermes. Given that each modality encompasses a significant amount of variation, we find that a modality prior of length 1 does not possess adequate capacity to encode modality-related information. We observe that longer modality tokens further improve the performance, but the gains saturate as the length increases. Therefore, we choose $l=10$ for our main experiments.

\begin{table}[h]
    \centering
    \caption{Ablation study on the length of modality prior token. We report the average Dice on the seven datasets with Hermes implemented with the ResUNet-Small backbone. }
    \label{tab:appendix_modality_length}
         \begin{tabular}{ccccc} 
         \toprule
          $l$ & 1 & 5 & 10 & 20 \\
          \midrule
          Avg Dice & 82.93 & 83.16 & 83.37 & 83.38 \\
          \bottomrule
      \end{tabular}

\end{table}

\begin{figure}
    \centering
    \includegraphics[width=0.46\textwidth]{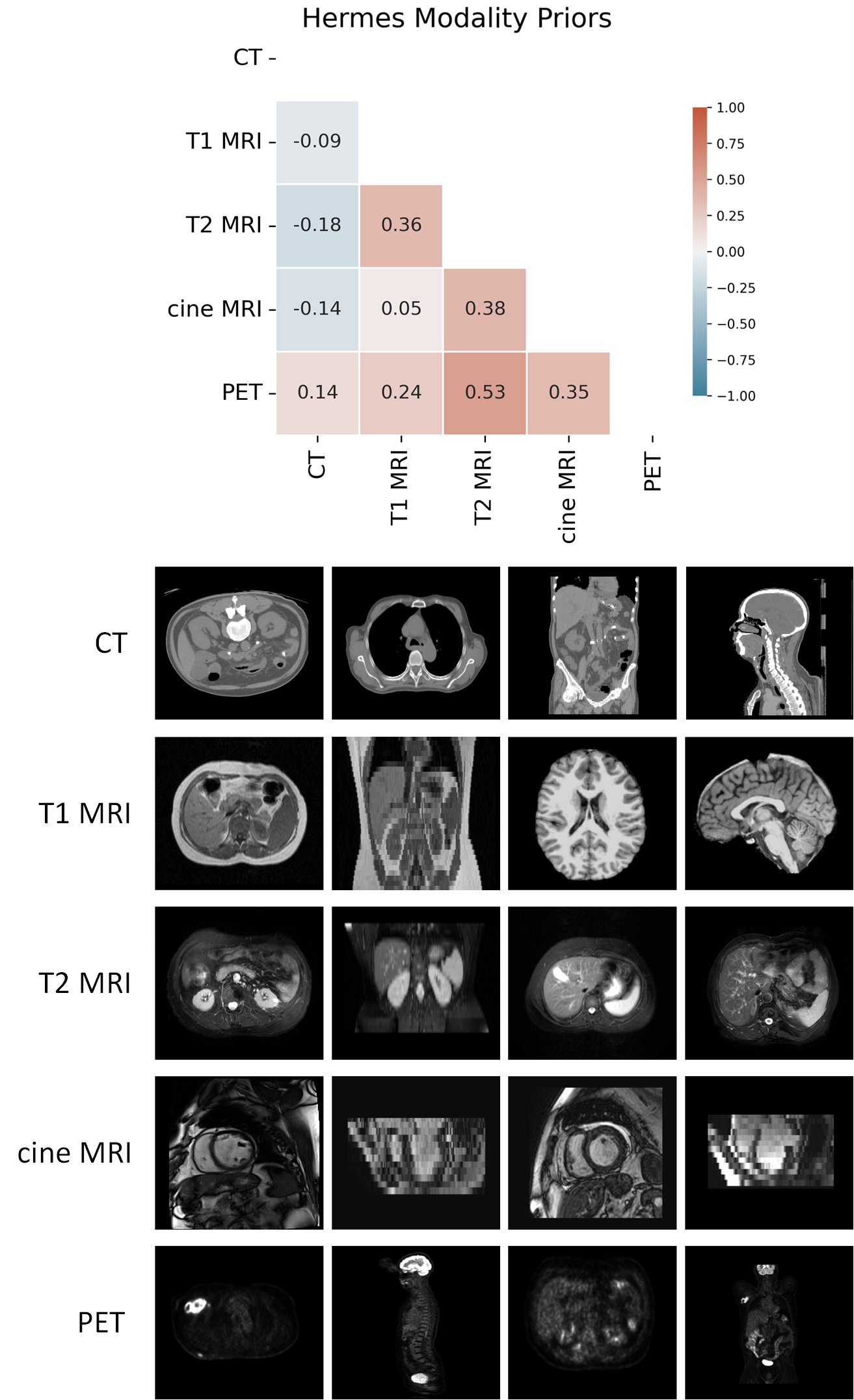}
    \caption{Upper: The cosine similarity between the learned Hermes modality priors. Lower: Illustration of each imaging modality.}
    \label{fig:sup_modality_priors}
\end{figure}

\textbf{Additional analysis on the learned priors.} In this section, we briefly introduce the imaging principle and the typical visual appearance of structures in different modalities to help interpret Hermes's learned modality priors, see Fig. \ref{fig:sup_modality_priors}.

CT uses X-ray beams to create detailed cross-sectional images of the body. In CT images, bones and other dense structures appear very bright (high attenuation), while soft tissues show up in varying shades of gray. Air and other gas-filled spaces appear dark due to their low X-ray absorption. CT is particularly effective for visualizing bones, lung tissue, and detecting abnormalities like tumors or fractures.

T1-weighted MRI utilizes magnetic fields and radiofrequency pulses to produce detailed images of the body's internal structures. In T1-weighted images, fat-containing tissues appear bright, and water-rich tissues look darker. This contrast makes T1 MRI particularly useful for visualizing fine anatomical details, such as the brain's white and gray matter.

T2-weighted MRI also uses magnetic fields and radio waves but with different timing parameters than T1, leading to different tissue contrasts. In T2 images, fluid-containing tissues appear bright, while fat appears darker compared to T1 images. This characteristic makes T2 MRI ideal for visualizing fluid-filled spaces and edema. 

Cine MRI is a specialized form of MRI used to capture moving images of the body. It is particularly useful in cardiology for visualizing the heart's movement and blood flow. In cine MRI, fluid dynamics are emphasized, making it excellent for assessing cardiac function, valve abnormalities, and congenital heart disease. The visualization of blood flow and moving structures is a unique aspect of cine MRI.

PET (Positron Emission Tomography) scans use radioactive tracers to detect metabolic activities within the body. The patient is injected with a radiotracer, which accumulates in areas of high metabolic activity. PET scanners detect the gamma rays emitted by the tracer and use this data to construct images. In PET images, areas of high tracer uptake, such as rapidly growing cancer cells, appear brighter. PET is highly effective in cancer diagnosis, as it can reveal the metabolic activity of tumors.

In Fig. \ref{fig:sup_modality_priors}, we present the cosine similarity between the modality priors as learned by Hermes. Consistent with imaging principles, Hermes identifies CT as distinctly different from MRI and PET modalities, owing to its unique X-ray based imaging technique and contrasting tissue visualization. This finding aligns with the principle that CT images bone and dense structures more effectively, setting it apart from MRI and PET techniques.

Regarding MRI sequences, Hermes notes a higher similarity among them compared to CT, reflecting their shared basis in magnetic resonance techniques, even though they provide different tissue contrasts. Specifically, Hermes discerns a closer relationship between cine MRI and T2 MRI, attributable to their shared emphasis on fluid content visualization. This is in line with cine MRI's application in capturing moving structures like blood flow, similar to the fluid-highlighting characteristics of T2 MRI images.

Additionally, Hermes finds PET imaging to be more similar to T2 and cine MRI than to T1 MRI. This observation can be understood through PET's focus on metabolic activities and functional changes, aspects that are also emphasized to some extent in T2 and cine MRI, despite their different fundamental principles. On the other hand, the lower similarity with T1 MRI is logical, given T1 MRI's distinct imaging focus, primarily on fat visualization, as opposed to PET's metabolic activity emphasis.

These findings by Hermes are in accordance with the fundamental principles of these imaging modalities. They indicate that Hermes is effectively capturing the unique imaging characteristics and tissue contrasts inherent to each modality, demonstrating a sophisticated understanding of how different imaging techniques visualize various tissues and physiological processes.

\section{Dataset details}
In this section, we provide detailed information about each dataset, including the volume of data, annotation categories, data sources, as well as how we use them for training and testing. At last, we explain how we designed our experiments using these datasets.

\begin{table}
  \caption{Datasets statistics. The upper datasets are for upstream training and analysis. The bottom two datasets are for downstream tasks on transfer learning, incremental learning, and generalization.}
  \label{tab:dataset}
  \centering
   \scriptsize
   \setlength{\tabcolsep}{1mm}{
  \begin{tabular}{ccccccc}
    \toprule
    Dataset         & Body Region   & Modality  & Clinical Target   & \#Cls &   Size     \\
    \midrule
    BCV~\cite{bcv}     & Abdomen       & CT        & Organs            & 13        & 30    \\
    LiTS~\cite{bilic2019liver}     & Abdomen       & CT        & Liver \& Tumor    & 2         & 131   \\
    KiTS~\cite{heller2019kits19}     & Abdomen       & CT        & Kidney \& Tumor    & 2         & 210   \\
    AMOS CT~\cite{ji2022amos}  & Abdomen       & CT        & Organs            & 15        & 300   \\
    SS T\cite{structseg}& Thorax        & CT        & Organs            & 6         & 50    \\
    SS H \cite{structseg} & Head \& Neck & CT   & Organs & 22 & 50\\
    AMOS MR~\cite{ji2022amos}  & Abdomen       & MRI       & Organs            & 13        & 60    \\
    CHAOS~\cite{CHAOS2021}    & Abdomen       & T1 \& T2 MRI & Organs         & 4         & 60    \\
    M\&Ms~\cite{campello2021multi} & Cardiac & cineMRI & Structures & 3 & 320 \\
    DLBS~\cite{rodrigue2012beta} & Brain & T1 MRI & Structures & 3 & 213 \\
    AutoPET~\cite{gatidis2022whole} & Whole body & PET & Lesions & 1 & 1014 \\
    \midrule
    SegTHOR~\cite{lambert2020segthor}  & Thorax        & CT        & Organs            & 3         & 40    \\
    MSD Pancreas~\cite{antonelli2022medical}& Abdomen       & CT        & Pancreas \& Tumor    & 2         & 281 \\

    \bottomrule
  \end{tabular}}
\end{table}

\textbf{BCV dataset.} The BCV dataset~\cite{bcv} (Multi-Atlas Labeling Beyond the Cranial Vault) comprises 50 subjects with abdominal CT scans, of which 30 training images are publicly available. Thirteen abdominal organs were manually labeled on a volumetric basis using the MIPAV software. The labeled organs include the spleen, right kidney, left kidney, gallbladder, esophagus, liver, stomach, aorta, inferior vena cava, portal vein and splenic vein, pancreas, right adrenal gland, and left adrenal gland. Some patients may lack the right kidney or gallbladder, and therefore these organs are not labeled. All scans were acquired for routine clinical care from CT scanners at the Vanderbilt University Medical Center (VUMC). The BCV dataset is used as one of the seven datasets for upstream training. We randomly select 75\%/5\%/20\% of the publicly available images for training/validation/testing.

\textbf{LiTS dataset.} 
The LiTS dataset~\cite{bilic2019liver} (Liver Tumor Segmentation Challenge) comprises 201 computed tomography (CT) images of the abdomen, with 131 training cases and 70 testing cases, where only the label of training cases are publicly available. The LiTS dataset provides detailed annotation for tumors while offering coarse annotation for the liver. The image data originates from various clinical sites, including Ludwig Maxmilian University of Munich, Radboud University Medical Center of Nijmegen, Polytechnique \& CHUM Research Center Montréal, Tel Aviv University, Sheba Medical Center, IRCAD Institute Strasbourg, and the Hebrew University of Jerusalem. The studied subjects suffer from diverse liver tumor diseases, such as hepatocellular carcinoma (HCC), as well as secondary liver tumors and metastases originating from colorectal, breast, and lung cancers. The tumors exhibit varying contrast enhancement, including hyper and hypo-dense contrast. The images represent a mix of pre- and post-therapy abdominal CT scans, acquired with different CT scanners and acquisition protocols. The LiTS dataset is used as one of the seven datasets for upstream training. We randomly select 75\%/5\%/20\% of the 131 training cases for training/validation/testing.

\textbf{KiTS dataset.} 
The KiTS19 dataset~\cite{heller2019kits19} comprises segmented CT imaging and treatment outcomes for 300 patients who underwent partial or radical nephrectomy for one or more kidney tumors at the University of Minnesota Medical Center between 2010 and 2018. Out of these cases, 210 have been released publicly, while the remaining 90 are kept private for evaluation purposes.  The KiTS is used as one of the seven datasets for upstream training. We randomly select 75\%/5\%/20\% of the 210 training cases for training/validation/testing.

\textbf{AMOS CT \& MR dataset.}
The AMOS dataset~\cite{ji2022amos} is a large-scale collection of CT and MRI data from 600 patients diagnosed with abdominal tumors or abnormalities at Longgang District People's Hospital. The dataset comprises 500 CT and 100 MRI scans acquired from eight different scanners and vendors, encompassing 15 organ categories: spleen, right kidney, left kidney, gallbladder, esophagus, liver, stomach, aorta, inferior vena cava, pancreas, right adrenal gland, left adrenal gland, duodenum, bladder, and prostate/uterus. For CT images, AMOS provides 200 scans for training and 100 scans for validation, while for MRI images, 40 scans are provided for training and 20 scans for validation. Both AMOS CT and AMOS MR are used as two of the seven datasets for upstream training. In line with the AMOS benchmark paper \cite{ji2022amos}, we report testing performance on the official validation set and utilize 95\%/5\% training data for model training/validation. As all images in the AMOS MR validation set don't have the annotation of bladder and prostate, we only segment 13 organs for AMOS MR.

\textbf{StructSeg dataset.}
The StructSeg dataset~\cite{structseg} is collected from a challenge for the segmentation of organs-at-risk (OAR) and gross target volume (GTV) of tumors of two types of cancers, nasopharynx cancer and lung cancer, for radiation therapy planning. We use Task 1 and 3, organ-at-risk segmentation from head\&neck and thorax CT scans in our experiments, denoted as SS H and SS T respectively. SS H has 22 OAR annotations from 50 nasopharynx cancer patients, including left eye, right eye, left lens, right lens, left optical nerve, right optical nerve, optical chiasma, pituitary, brain stem, left temporal lobes, right temporal lobes, spinal cord, left parotid gland, right parotid gland, left inner ear, right inner ear, left middle ear, right middle ear, left temporomandibular joint, right temporomandibular joint, left mandible and right mandible. SS T has 6 OARs annotated on CT scans from 50 lung cancer patients, including left lung, right lung, spinal cord, esophagus, heart, and trachea. We split the scans into 75\%/5\%/20\% for training/validation/testing.

\textbf{CHAOS dataset.}
The CHAOS dataset~\cite{CHAOS2021} is collected from a challenge for the precise segmentation of abdominal organs. We use the data from Task 5: segmentation of abdominal organs from MRI. Four organs, including the liver, left kidney, right kidney, and spleen are annotated. They provide three MR sequences, including T1-in-phase, T1-out-phase, and T2-SPIR, for 20 patients. We treat different MR sequences as separate images and split the dataset at the patient level into 75\%/5\%/20\% for training/validation/testing.

\textbf{M\&Ms dataset.} 
The M\&Ms dataset~\cite{campello2021multi} is from Multi-Centre, Multi-Vendor and Multi-Disease Cardiac Segmentation (M\&Ms) Challenge, which was organized as part of the MICCAI 2020 Conference. This dataset cohort includes patients with hypertrophic and dilated cardiomyopathies and healthy subjects. All subjects were scanned in clinical centres in three different countries (Spain, Germany, and Canada) using four different MRI scanner vendors (Siemens, General Electric, Philips, and Canon). The training set contains 150 annotated images from two vendors (75 each), while the testing set contains 170 cases (20 for the first vendor and 50 each for the other three vendors). Three categories of annotation are 61 provided at the end-diastolic (ED) and end-systolic (ES) 62 phase, including left ventricle (LV), right ventricle (RV), and 63 left ventricular myocardium (MYO). We use the official testing set for testing, and divide the training set into 95\% for training and 5\% for validation.

\textbf{DLBS dataset.}
The Dallas Lifespan Brain Study (DLBS)~\cite{rodrigue2012beta} is designed to understand the antecedents of preservation and decline of cognitive function at different stages of the adult lifespan, with a particular interest in the early stages of a healthy brain’s march towards Alzheimer Disease. We use the 213 T1 MRI scans to segment the cerebrospinal fluid, gray matter and white matter. Following~\cite{rao2022improving}, we divide the 213 scans into 129 for training, 43 for validation, and 43 for testing.

\textbf{AutoPET dataset.}
The AutoPET dataset~\cite{gatidis2022whole} provides annotated Positron Emission Tomography/Computed Tomography (PET/CT) studies, encompassing a significant collection of 1014 whole-body Fluorodeoxyglucose (FDG)-PET/CT datasets. This dataset includes 501 studies from patients diagnosed with malignant lymphoma, melanoma, and non-small cell lung cancer (NSCLC), alongside 513 studies serving as negative controls without PET-positive malignant lesions. We divide the dataset at the patient level into 75\%/5\%/20\% for train/validation/testing.

\textbf{SegTHOR dataset (SS T and SS H).}
The SegTHOR dataset~\cite{lambert2020segthor} aims at the thoracic organ-at-risk segmentation in CT images. This dataset provides 4 OARs annotations from 40 CT scans, including heart, aorta, trachea, and esophagus. We use the SegTHOR dataset as a downstream task to evaluate the generalization of models.  We directly use the upstream-trained model to make predictions on all 40 images and report the generalization performance.

\textbf{MSD pancreas \& tumor dataset.}
The MSD pancreas \& tumor dataset is a part of the Medical Image Segmentation Decathlon (MSD)~\cite{antonelli2022medical}, an international challenge aimed at identifying a general-purpose algorithm for medical image segmentation. The competition encompasses ten distinct datasets featuring various target regions, modalities, and challenging attributes. MSD pancreas \& tumor is one of the datasets that is annotated for pancreas and tumors. The shape and position of tumors vary greatly between patients. The MSD pancreas \& tumor dataset consists of 281 CT images. We use it as a downstream task to evaluate models' capacity for transfer learning and incremental learning. We split the dataset into 214 samples for training, 10 samples for validation, and 57 samples for testing. To evaluate the impact of downstream data volume, we conducted experiments on 1\%, 10\%, 50\%, and 100\% of the 214 training samples. To reduce the variability from the training sample selection, we report the average performance over 5 runs for the 1\% and 10\% settings.

\textbf{Experiment design.}
To substantiate the efficacy of the proposed universal medical image segmentation paradigm, we have meticulously curated these datasets, see Table \ref{tab:dataset}. These datasets were selected based on three main factors: anatomical regions, imaging modalities, and clinical targets. The careful selection of these upstream training datasets is designed to provide comprehensive answers to the three research questions originally posed in our introduction section. For the downstream tasks, we chose the challenging MSD pancreas \& tumor dataset for transfer learning and incremental learning. The pancreas is a relatively small, elongated glandular organ, while the shape and location of a pancreatic tumor can greatly vary. As such, the segmentation difficulty of this task is extremely high. Furthermore, this dataset is comprised of a large number of images, with 281 CT scans, allowing us to adequately test the model's transfer learning and incremental learning abilities under various downstream data volumes. In addition, we select the SegTHOR dataset to verify the model's generalization performance. There is only one thoracic dataset (StructSeg) in the upstream training. The StructSeg and SegTHOR are both for thoracic OAR segmentation and have three overlap categories of heart, trachea, and esophagus. Evaluating performance on these overlapping categories allows us to explore the universal paradigm's potential generalization ability to different anatomical regions and analyze whether more abdominal tasks contribute positively to the generalization of thoracic tasks.

\end{document}